\documentclass{article}

\usepackage[preprint]{neurips_2026}

\usepackage[utf8]{inputenc} 
\usepackage[T1]{fontenc}    
\usepackage{graphicx}       
\usepackage{float}          
\usepackage{hyperref}       
\usepackage{url}            
\usepackage{booktabs}       
\usepackage{amsmath}        
\usepackage{amsfonts}       
\usepackage{nicefrac}       
\usepackage{microtype}      
\usepackage{multirow}       
\usepackage{colortbl}       
\usepackage{xcolor}         

\title{MedMIX: Modality-Internal Expert Fusion for Multimodal Medical Diagnosis}

 \author{%
   Seungik Cho\thanks{Equal contribution.} \\
   Department of Physics and Astronomy\\
   Rice University\\
   Houston, TX 77005 \\
   \texttt{sc252@rice.edu} \\
   \And
   Anqi Li\footnotemark[1] \\
   Department of Electrical and Computer Engineering\\
   Rice University\\
   Houston, TX 77005 \\
   \texttt{al318@rice.edu} \\
   \AND
   Wei Qiu\thanks{Corresponding author. \texttt{wq8@rice.edu}} \\
   Department of Electrical and Computer Engineering\\
   Rice University\\
   Houston, TX 77005 \\
   \texttt{wq8@rice.edu} \\
 }

\begin{document}

\maketitle

\begin{abstract}
Multimodal clinical prediction faces three challenges: multiple foundation models (FMs) with complementary strengths per modality, pervasive missing modalities at training and test time, and sample-specific variation in modality contributions. We introduce \textbf{MedMIX}, a multimodal framework that combines intra-modality expert fusion, learned inter-modality fusion, and training-only large--small model collaboration for robust medical prediction under incomplete modalities. Within each modality, MedMIX aggregates complementary embeddings from multiple small expert models; across modalities, it performs learned fusion over available modalities; and during training, it leverages large teacher models to improve deployed representations without additional inference cost. Across three heterogeneous benchmarks (OpenI, MIMIC-IV-MM, and MMIST-ccRCC), MedMIX achieves consistently strong performance while remaining robust under controlled missing-modality perturbations, and further demonstrates sustained robustness under cross-cohort shift on MIMIC-III. These results highlight MedMIX as a practical framework that unifies within-modality expert collaboration, sample-specific cross-modality fusion, and efficient large--small model collaboration while remaining robust to incomplete modalities. Code can be found at \url{https://anonymous.4open.science/r/MedMIX-8DD4/}.

\end{abstract}

\section{Introduction}


Multimodal clinical prediction increasingly relies on heterogeneous healthcare data, such as medical images, clinical text, and clinical time series~\citep{acosta2022multimodal,soenksen2022integrated}. Foundation models have expanded the pool of reusable feature extractors for these modalities, creating a heterogeneous model ecosystem with diverse representation sources, training paradigms, and capacity scales. These models can provide complementary views of the same clinical sample, but their diversity also makes effective collaboration across models and modalities increasingly important in practice. These challenges manifest along multiple dimensions, including model diversity, data incompleteness, and variability in modality relevance across samples.

\paragraph{Challenge \textcircled{1}: Multiple complementary experts per modality, no principled aggregation.}
For any single modality, a practitioner faces not one but several viable FMs, often spanning a range of model scales and deployment settings and arising from different training setups and data sources. These models often exhibit distinct and complementary strengths. A general-purpose model may capture broad contextual representations~\citep{oquab2023dinov2, qwen2}, whereas a domain-specific biomedical model captures clinically salient patterns~\citep{yang2022gatortron, zhang2023biomedclip}, and retrieval-augmented models ground representations in external knowledge sources such as literature~\citep{lewis2020retrieval, jin2023medcpt}. In addition, practitioners often need to leverage both lightweight deployable models and larger high-capacity models, further complicating how complementary expert representations can be effectively integrated in practice~\citep{hoffmann2022training, kaplan2020scaling, chen2025multi}. Existing biomedical multimodal frameworks either select a single backbone per modality, thereby discarding complementary information and reducing robustness to distributional shift, or naively concatenate all embeddings, ignoring inter-expert redundancy and the potentially large variance in embedding dimensionality across models~\citep{tripathi2025honeybee, chen2025multi}. Consequently, there is a lack of a principled formulation for aggregating multiple modality-specific experts within a single modality prior to cross-modal fusion.

\paragraph{Challenge \textcircled{2}: Pervasive and structured missing modalities.}
Clinical data collection in real-world healthcare settings is inherently opportunistic: a patient admitted to the ICU may lack imaging data if imaging is not clinically indicated, or have imaging available but no pathology data if no biopsy has been performed. Studies report that between 30\% and 70\% of samples in real-world cohorts have at least one missing modality~\citep{zhang2022m3care}. Critically, missingness is often not purely random: it correlates with disease severity, clinical protocol, and socioeconomic factors, creating structured patterns that models must handle at both train and test time. Strategies that simply zero-impute missing embeddings or naively skip missing modalities without explicitly modeling missingness can conflate missingness with weak signal and degrade performance as missingness rates increase~\citep{wu2024multimodal}.

\paragraph{Challenge \textcircled{3}: Sample-specific modality contributions.}
Even when all modalities are nominally present, their contribution to the final prediction  can vary substantially from sample to sample. For example, one modality may contain the dominant clinical evidence for a given case, whereas another may provide weak, redundant, or conflicting information. Uniform fusion can therefore underuse informative modalities or overemphasize less relevant inputs. Prior work on trusted multi-view learning~\citep{han2022trusted} addresses this at the feature level, but does not directly address the combined setting of within-modality expert aggregation, missing modalities, and logits-level cross-modality fusion.

\begin{figure}[t]
    \centering
    \includegraphics[width=\textwidth]{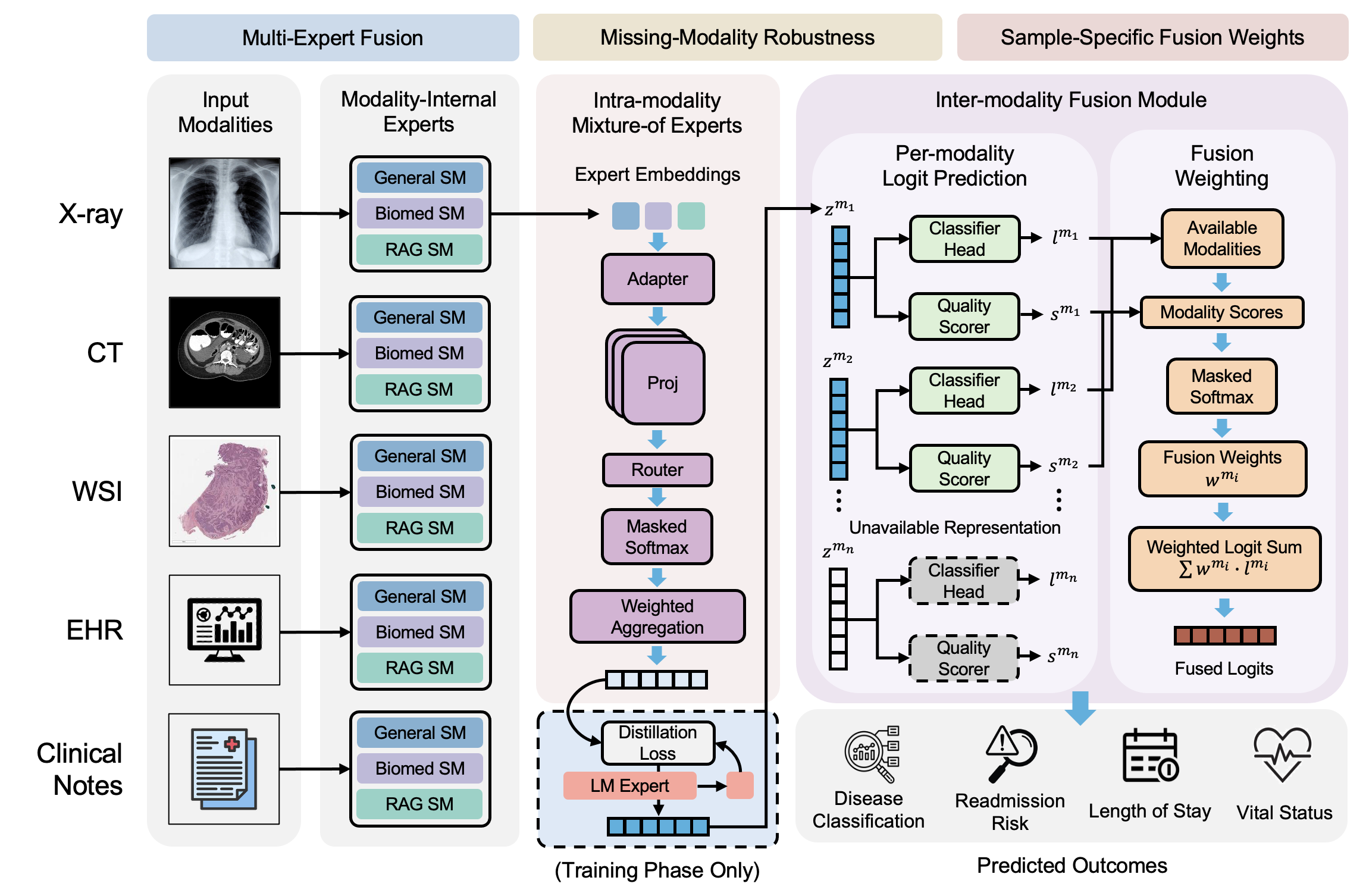}
    \caption{Overview of MedMIX. Within each modality, multiple expert encoders from small models (SMs) are aggregated through a learned MoE router to obtain a unified representation. Across modalities, per-modality logits are fused using sample-specific weights, with missing modalities masked out. A training-only distillation objective further aligns SM representations with larger teacher models (LMs).}
    \label{fig:medmix_overview}
\end{figure}

To address these three challenges simultaneously, we propose \textbf{MedMIX} (\textbf{Med}ical \textbf{M}odality-\textbf{I}nternal e\textbf{X}pert fusion; Figure \ref{fig:medmix_overview}), a unified framework for robust multimodal clinical prediction that combines lightweight expert models with training-time guidance from large teacher models. MedMIX is organized into two stages of fusion. \emph{Intra-modality}: for each modality, lightweight residual adapters refine each expert's embedding into a shared latent space; a learned MoE router then computes soft aggregation weights over available experts, adapting to heterogeneous experts while masking missing ones. \emph{Inter-modality}: each modality produces per-sample logits via an independent classifier head, and a learned fusion scorer assigns sample-specific weights across available modalities, capturing varying modality contributions, while missing modalities are excluded from the softmax normalization, so fusion weights are learned and re-normalized over the observed modalities. An optional training-only distillation objective aligns lightweight representations with large teacher models, improving sample efficiency without incurring inference overhead.

We evaluate MedMIX on four benchmark datasets spanning chest pathology classification (OpenI), ICU outcome prediction (MIMIC-IV-MM), cancer survival prediction (MMIST-ccRCC), and cross-cohort generalization (MIMIC-III). Across benchmarks and controlled missing-modality settings, MedMIX consistently achieves strong performance while maintaining robustness to incomplete modalities, with a favorable efficiency--accuracy trade-off.

Our main contributions are as follows.
\begin{itemize}
  \item \textbf{MedMIX}, a two-level fusion framework that addresses intra-modality expert diversity, sample-specific inter-modality weighting, and training-time large--small model collaboration within a unified architecture. To our knowledge, this is among the first works to explicitly model the aggregation of complementary foundation-model experts within each modality for downstream clinical prediction.

  \item \textbf{Comprehensive missing-modality robustness evaluation.} We conduct systematic controlled missing-modality sweeps, test-time random-missing-rate sweeps, and external cross-cohort validation on MIMIC-III, showing that MedMIX maintains strong performance under incomplete-modality conditions.

  \item \textbf{Strong efficiency--accuracy trade-off.} Across the evaluated benchmarks, MedMIX achieves strong predictive performance while retaining a lightweight inference pathway, yielding a favorable efficiency--accuracy trade-off among compared multimodal methods.
\end{itemize}

\section{Related Work}

\paragraph{Foundation models in healthcare.}
The landscape of biomedical FMs has expanded rapidly across three axes:
vision, language, and temporal data. In medical imaging, models such as
MedSAM~\citep{ma2024segment} for segmentation, Prov-GigaPath~\citep{xu2024whole}
for whole-slide pathology, and Merlin~\citep{blankemeier2024merlin} for 3D CT
bring strong representations trained on millions of annotated images. In clinical
language, the trajectory from encoder-based models (BioBERT~\citep{lee2020biobert},
GatorTron~\citep{yang2022gatortron}) to decoder-based architectures
(BioMistral~\citep{labrak2024biomistral}, Me-LLaMA~\citep{xie2024me}) has produced
increasingly rich text encoders; Llm2Vec~\citep{behnamghader2024llm2vec} further
converts autoregressive LLMs into dense encoders. For temporal physiological signals,
MOIRAI~\citep{woo2024moirai} and Chronos~\citep{ansari2024chronos} provide
universal time-series representations pretrained across diverse domains.
Despite these advances, the models above are predominantly single-modal:
extending them directly to multimodal fusion remains non-trivial, motivating
the need for frameworks that can orchestrate multiple such specialists.

\paragraph{Multimodal medical prediction and fusion.}
Traditional fusion strategies operate at three levels—data, feature, and
decision~\citep{ramachandram2017deep}—and have been instantiated through
transformer-based cross-modal attention~\citep{zhou2023transformer},
graph-based entity modeling~\citep{zheng2022multi}, and shared latent
space learning~\citep{soenksen2022integrated}.
More recently, general-purpose multimodal encoders (Meta-Transformer~\citep{zhang2023meta},
OneLLM~\citep{han2024onellm}, AnyGPT~\citep{zhan2024anygpt}) have been fine-tuned on clinical tasks,
but they struggle on small multimodal medical datasets and impose high computational costs.
A complementary line of work explores collaboration between large and small models:
large models supply rich representations or supervision signals, while small models handle
efficient deployment~\citep{hinton2015distilling,xu2024small,zhang2018dml,guo2020online}.
AdaCoMed~\citep{chen2025multi} is the most directly related prior work,
combining a Mixture-of-Modality-Experts fusion of single-modal large models with a
multimodal small model via contrastive alignment and adaptive weighting.
However, it applies contrastive-similarity weighting rather than a learned fusion rule.

\paragraph{Missing modality learning.}
Prior approaches fall into three broad categories: imputation-based methods that reconstruct missing views from observed ones~\citep{tran2017missing,zhang2022m3care}, Bayesian or probabilistic formulations that model missingness explicitly~\citep{ma2021smil}, and representation-learning strategies such as masked autoencoders and modality-dropout curricula that build robustness into the encoder~\citep{wu2024multimodal,wu2026remind}. MedMIX adopts a simpler, complementary strategy: rather than imputing, it explicitly zeros out unavailable expert or modality representations and uses masked softmax to re-normalize fusion weights accordingly. This avoids hallucinating absent signal and remains easy to implement in practice.

\paragraph{Mixture-of-experts architectures.}
Sparse MoE models~\citep{shazeer2017outrageously,fedus2022switch} increase network
capacity without proportional inference cost and have been widely used in
multi-task learning, multilingual NLP, and multimodal settings~\citep{mustafa2022multimodal}.
In medical imaging, MoE layers have been used to specialize sub-networks by anatomy
or imaging modality~\citep{chen2025multi}. MedMIX adopts the MoE paradigm at a
finer granularity: instead of routing tokens or samples to expert \emph{networks},
it assigns query-specific soft weights to a fixed committee of pre-extracted expert embeddings within each modality, functioning as a dense aggregation rather than conditional sparse computation. This design enables asynchronous updates
as new FMs emerge, without retraining the fusion module from scratch.

\paragraph{Knowledge distillation for medical representation learning.}
Knowledge distillation~\citep{hinton2015distilling} transfers supervision from large teacher models to compact student networks, and has been applied in medical AI to improve representation quality under deployment constraints~\citep{liu2023distilling}. MedMIX employs a training-only objective combining cosine alignment and relational knowledge distillation~\citep{park2019relational}, discarding teacher encoders at inference time to improve deployed representations without additional inference cost.

\section{Method}

\subsection{Preliminary}
Given an input sample with $M$ modalities, each modality $m$ is represented by $K_m$ pre-extracted expert embeddings $\{e_k^{(m)}\}_{k=1}^{K_m}$ from lightweight (small) foundation models (SMs) (biomedical, general-purpose, and retrieval-augmented). Here the retrieval-augmented branch is treated as another expert within the same modality, i.e., as a way to instantiate within-modality model diversity rather than as an additional modality. As illustrated in Figure~\ref{fig:medmix_overview}, our model first aggregates experts within each modality via \emph{Intra-Modality MoE}, then produces per-modality logits using independent classifier heads, and finally combines them through \emph{learned logits-level inter-modality fusion} with mask-aware re-normalization over available modalities. During training only, an optional teacher--student distillation objective aligns modality representations with larger (large) teacher models (LMs).

\subsection{Expert Committee Design}
Rather than relying on a single FM backbone, each modality in MedMIX is processed by a committee of three complementary expert types. \emph{General} experts are pretrained on broad corpora and supply transferable, wide-coverage representations. \emph{Biomedical} experts are pretrained on domain-curated data and capture pathology-relevant semantics that general models underweight. \emph{RAG} experts augment representations with external biomedical knowledge retrieved from structured or unstructured sources, providing evidence beyond what parametric encoders encode. These three axes---breadth, domain depth, and retrieval-augmented grounding---are designed to be complementary, yielding the greatest diversity particularly under distribution shift or ambiguous inputs where ensemble disagreement is most informative. Full details of the RAG embedding construction are in \ref{sec:appendix_rag}.

\subsection{Intra-Modality Expert Aggregation}
 
Each expert $k$ in modality $m$ is first refined by a lightweight residual adapter:
\begin{equation}
\begin{aligned}
\tilde{e}_{k}^{(m)} 
&= e_{k}^{(m)} + \mathrm{Adapter}_{k}^{(m)}\bigl(e_{k}^{(m)}\bigr), 
\mathrm{Adapter}_{k}^{(m)} &:\; d_k^{(m)} \to r_k^{(m)} \to d_k^{(m)}.
\end{aligned}
\end{equation}
where $r_k^{(m)} = \min(128, \max(32, \lfloor d_k^{(m)}/16 \rfloor))$. We then project each refined embedding to a shared space as
$z_k^{(m)} = \mathrm{Proj}_k^{(m)}\bigl(\tilde{e}_k^{(m)}\bigr) \in \mathbb{R}^{d}$ using a Linear--LayerNorm--GELU--Dropout block.
Missing experts are masked by zeroing: $z_k^{(m)} \leftarrow z_k^{(m)} \cdot \mathbf{1}[\mathrm{mask}_k^{(m)}]$, where $\mathrm{mask}_k^{(m)} = 1$ indicates that expert $k$ of modality $m$ is present and $0$ otherwise.
 
Routing uses a modality-specific shared per-expert scorer $s_k^{(m)} = \mathrm{Router}^{(m)}(z_k^{(m)}) \in \mathbb{R}$. Unavailable experts are masked to $-10^4$ before the softmax:
\begin{equation}
\begin{aligned}
\boldsymbol{g}^{(m)} 
&= \mathrm{softmax}\!\bigl(\mathbf{s}^{(m)}\bigr), 
z^{(m)} &= \Bigl(\sum_k g_k^{(m)} z_k^{(m)}\Bigr)\cdot
\mathbf{1}\!\Bigl[\textstyle\sum_k \mathrm{mask}_k^{(m)} > 0\Bigr].
\end{aligned}
\end{equation}
 
\subsection{Learned Inter-Modality Fusion}
 
Each modality representation $z^{(m)}$ is decoded into per-modality logits $\ell^{(m)} \in \mathbb{R}^C$ through an independent classifier head. A learned fusion score $s^{(m)} = \mathrm{Scorer}^{(m)}(z^{(m)})$ provides a sample-specific weighting signal for modality $m$. Unavailable modalities are assigned a large negative score ($\tilde{s}^{(m)} = -10^4$) before the softmax, so fusion weights are re-normalized over the observed modality set via masked softmax. The fused logits are
\begin{equation}
    w^{(m)} = \mathrm{softmax}(\tilde{\mathbf{s}})^{(m)}, \qquad
    \ell_{\mathrm{fused}} = \sum_m w^{(m)} \ell^{(m)},
\end{equation}
a learned weighted combination of available modality logits. A task-specific activation (sigmoid for multi-label classification, softmax for multi-class) is applied to $\ell_{\mathrm{fused}}$ to produce the final prediction. The scorer is trained jointly with the task objective, so fusion weights adapt end-to-end to both modality content and availability.
 
\subsection{Teacher–Student Distillation}

To strengthen lightweight representations without increasing inference cost, we add a training-only distillation objective.
For each modality $m$, a frozen large teacher model produces embedding $z_T^{(m)}$, projected to dimension $d$ via a learned linear head.
Two alignment terms are used for each modality: cosine distillation,
$\mathcal{L}_\mathrm{cos}^{(m)} = 1 - \cos\bigl(z^{(m)}, z_T^{(m)}\bigr)$,
and relational knowledge distillation (RKD), denoted by
$\mathcal{L}_\mathrm{RKD}^{(m)}$, which aligns pairwise distance structures among available samples in the batch.
We define the per-modality distillation objective as
$\mathcal{L}_\mathrm{distill}^{(m)} = \mathcal{L}_\mathrm{cos}^{(m)} + \lambda_\mathrm{RKD}\,\mathcal{L}_\mathrm{RKD}^{(m)}$,
and the total distillation loss as
$\mathcal{L}_\mathrm{distill} = \sum_m a^{(m)}\mathcal{L}_\mathrm{distill}^{(m)}$,
where $a^{(m)}\in\{0,1\}$ indicates whether modality $m$ is available for the current sample.
The distillation weight ramps from 0 to $\lambda_D^\mathrm{max}$ over $T_D$ warmup epochs.
The total loss is:
\begin{equation}
    \mathcal{L} = \mathcal{L}_\mathrm{task} + \lambda_D(t)\,\mathcal{L}_\mathrm{distill}.
\end{equation}

\section{Results}
\subsection{Experimental Setup}
\noindent\textbf{Datasets.}
We evaluate MedMIX on four multimodal medical datasets.
\textbf{OpenI}~\citep{demner2016openi} contains chest X-rays paired with radiology reports. The task is multi-label disease classification.
\textbf{MIMIC-IV-MM}~\citep{johnson2023mimiciv,johnson2019mimiccxr,johnson2023mimicivnote} integrates chest X-rays, radiology notes, and time-series EHR data. We evaluate on three ICU outcome prediction tasks.
\textbf{MMIST-ccRCC}~\citep{mota2024mmist} provides CT, whole-slide images (WSI), and clinical variables from 618 ccRCC patients. The task is 12-month vital status prediction.
\textbf{MIMIC-III}~\citep{johnson2016mimiciii} serves as an external validation cohort to assess cross-cohort generalizability, using radiology reports and time-series EHR data.
Dataset splits, preprocessing, and modality details are in \ref{sec:appendix_reproducibility}.

\noindent\textbf{Baselines.}
We compare against simple fusion rules (MeanAvg, Concat, Max, Attention),
the missing-modality framework M3Care~\citep{zhang2022m3care},
the large--small collaboration framework AdaCoMed~\citep{chen2025multi},
and the unified multimodal encoder OneLLM~\citep{han2024onellm} fine-tuned on each task.
Modality-specific large teacher models (LMs) are also reported as single-modality references.
All embedding-based baselines share the same per-modality expert set as MedMIX, including the RAG branch.
Full baseline setup and expert assignments are in~\ref{sec:appendix_reproducibility}.

\noindent\textbf{Metrics and training.}
We report AUROC, AUPRC, macro-F1 (mF1), and Accuracy (Acc) as mean $\pm$ std over 5 seeds.
Implementation details are provided in~\ref{sec:appendix_reproducibility}.

\subsection{In-Domain Benchmark Results}
Table~\ref{tab:main_results} presents the main comparison on OpenI, MIMIC-IV-MM, and MMIST-ccRCC. In the table, entries labeled \texttt{*-Large} denote modality-specific teacher LMs used as single-modality reference baselines. Beyond M3Care, AdaCoMed, and OneLLM, we also compare against four simple fusion baselines, namely MeanAvg, Concat, Max, and Attention. MedMIX achieves the best performance on all four reported metrics on the in-domain benchmarks, outperforming both simple fusion baselines and the stronger multimodal baselines. Relative to the strongest baseline for each metric, the gains range from 0.6\%--4.0\% on OpenI, 1.8\%--9.1\% on MMIST-ccRCC, and 0.34\%--3.47\% on MIMIC-IV-MM. We therefore interpret the main comparison as evidence that modality-internal expert collaboration is beneficial on these three tasks, rather than as proof of universal superiority across all multimodal settings. Per-task results for the three MIMIC-IV-MM endpoints are reported separately in Appendix Table~\ref{tab:mimic_per_task}.
\begin{table}[h]
\centering
\caption{In-domain benchmark results on OpenI, MIMIC-IV-MM, and MMIST-ccRCC. Entries labeled \texttt{*-Large} denote modality-specific teacher large models used as single-modality reference baselines.}
\label{tab:main_results}
\resizebox{0.85\columnwidth}{!}{
\begin{tabular}{llcccc}
\toprule
Dataset & Method & AUROC & AUPRC & mF1 & Acc \\
\midrule
\multirow{11}{*}{OpenI}
& Front-Large & 0.7482 $\pm$ 0.0014 & 0.3155 $\pm$ 0.0037 & 0.3427 $\pm$ 0.0043 & 0.7626 $\pm$ 0.0211 \\
& Side-Large & 0.7104 $\pm$ 0.0059 & 0.2752 $\pm$ 0.0071 & 0.2977 $\pm$ 0.0058 & 0.7466 $\pm$ 0.0469 \\
& Note-Large & 0.9385 $\pm$ 0.0056 & \underline{0.7481 $\pm$ 0.0064} & 0.6980 $\pm$ 0.0098 & \underline{0.9363 $\pm$ 0.0034} \\
& MeanAvg & 0.8904 $\pm$ 0.0015 & 0.5485 $\pm$ 0.0091 & \underline{0.7201 $\pm$ 0.0040} & 0.8764 $\pm$ 0.0083 \\
& Concat & 0.8622 $\pm$ 0.0052 & 0.4817 $\pm$ 0.0185 & 0.6829 $\pm$ 0.0127 & 0.8557 $\pm$ 0.0114 \\
& Max & 0.8477 $\pm$ 0.0066 & 0.4566 $\pm$ 0.0049 & 0.6621 $\pm$ 0.0117 & 0.8361 $\pm$ 0.0162 \\
& Attention & 0.8477 $\pm$ 0.0035 & 0.4627 $\pm$ 0.0170 & 0.6563 $\pm$ 0.0140 & 0.8449 $\pm$ 0.0139 \\
& M3Care & \underline{0.9419 $\pm$ 0.0017} & 0.7270 $\pm$ 0.0059 & 0.6612 $\pm$ 0.0096 & 0.9265 $\pm$ 0.0051 \\
& AdaCoMed & 0.9260 $\pm$ 0.0027 & 0.6772 $\pm$ 0.0102 & 0.6297 $\pm$ 0.0151 & 0.9193 $\pm$ 0.0092 \\
& OneLLM & 0.8784 $\pm$ 0.0076 & 0.5635 $\pm$ 0.0230 & 0.7124 $\pm$ 0.0127 & 0.8732 $\pm$ 0.0173 \\
& MedMIX & \textbf{0.9570 $\pm$ 0.0006} & \textbf{0.7780 $\pm$ 0.0047} & \textbf{0.7246 $\pm$ 0.0061} & \textbf{0.9463 $\pm$ 0.0019} \\
\midrule
\multirow{11}{*}{MIMIC-IV-MM}
& XRay-Large & 0.6686 $\pm$ 0.0015 & 0.3900 $\pm$ 0.0011 & 0.4616 $\pm$ 0.0020 & 0.5248 $\pm$ 0.0116 \\
& Text-Large & 0.6810 $\pm$ 0.0032 & 0.4123 $\pm$ 0.0035 & 0.4696 $\pm$ 0.0030 & 0.5353 $\pm$ 0.0052 \\
& TimeSeries-Large & 0.5293 $\pm$ 0.0017 & 0.2689 $\pm$ 0.0013 & 0.4137 $\pm$ 0.0001 & 0.2997 $\pm$ 0.0003 \\
& MeanAvg & 0.7004 $\pm$ 0.0006 & 0.4147 $\pm$ 0.0017 & 0.6208 $\pm$ 0.0015 & 0.7111 $\pm$ 0.0120 \\
& Concat & 0.7008 $\pm$ 0.0013 & 0.4167 $\pm$ 0.0020 & 0.6166 $\pm$ 0.0047 & 0.7175 $\pm$ 0.0117 \\
& Max & 0.6866 $\pm$ 0.0022 & 0.3990 $\pm$ 0.0021 & 0.6021 $\pm$ 0.0046 & 0.6885 $\pm$ 0.0287 \\
& Attention & 0.6922 $\pm$ 0.0015 & 0.4012 $\pm$ 0.0038 & 0.6114 $\pm$ 0.0069 & 0.7108 $\pm$ 0.0307 \\
& M3Care & \underline{0.7068 $\pm$ 0.0017} & \underline{0.4435 $\pm$ 0.0018} & \underline{0.6319 $\pm$ 0.0053} & \underline{0.7251 $\pm$ 0.0368} \\
& AdaCoMed & 0.7001 $\pm$ 0.0013 & 0.4338 $\pm$ 0.0024 & 0.6275 $\pm$ 0.0025 & 0.7221 $\pm$ 0.0043 \\
& OneLLM & 0.5921 $\pm$ 0.0036 & 0.3344 $\pm$ 0.0073 & 0.5489 $\pm$ 0.0040 & 0.6681 $\pm$ 0.0145 \\
& MedMIX & \textbf{0.7168 $\pm$ 0.0019} & \textbf{0.4586 $\pm$ 0.0019} & \textbf{0.6375 $\pm$ 0.0035} & \textbf{0.7352 $\pm$ 0.0106} \\

\midrule
\multirow{11}{*}{MMIST-ccRCC}
& CT-Large & 0.6245 $\pm$ 0.0098 & 0.8811 $\pm$ 0.0086 & 0.4880 $\pm$ 0.0024 & 0.5521 $\pm$ 0.0033 \\
& WSI-Large & 0.5797 $\pm$ 0.0646 & 0.8945 $\pm$ 0.0201 & 0.5263 $\pm$ 0.0520 & 0.8331 $\pm$ 0.0500 \\
& Note-Large & 0.7372 $\pm$ 0.0077 & 0.9251 $\pm$ 0.0038 & \underline{0.7127 $\pm$ 0.0193} & \underline{0.8843 $\pm$ 0.0174} \\
& MeanAvg & 0.7075 $\pm$ 0.0064 & 0.9290 $\pm$ 0.0048 & 0.5867 $\pm$ 0.0814 & 0.7521 $\pm$ 0.1625 \\
& Concat & 0.7067 $\pm$ 0.0057 & 0.9367 $\pm$ 0.0017 & 0.6194 $\pm$ 0.0135 & 0.8628 $\pm$ 0.0193 \\
& Max & 0.6992 $\pm$ 0.0211 & 0.9285 $\pm$ 0.0100 & 0.5610 $\pm$ 0.0569 & 0.7306 $\pm$ 0.1411 \\
& Attention & 0.7113 $\pm$ 0.0174 & 0.9322 $\pm$ 0.0102 & 0.5917 $\pm$ 0.0636 & 0.8595 $\pm$ 0.0240 \\
& M3Care & 0.7024 $\pm$ 0.0127 & 0.9244 $\pm$ 0.0059 & 0.6207 $\pm$ 0.0262 & 0.8430 $\pm$ 0.0395 \\
& AdaCoMed & 0.7407 $\pm$ 0.0094 & \underline{0.9476 $\pm$ 0.0026} & 0.6813 $\pm$ 0.0097 & 0.8711 $\pm$ 0.0132 \\
& OneLLM & \underline{0.7445 $\pm$ 0.0131} & 0.9398 $\pm$ 0.0057 & 0.6983 $\pm$ 0.0042 & 0.8709 $\pm$ 0.0213 \\
& MedMIX & \textbf{0.8121 $\pm$ 0.0068} & \textbf{0.9673 $\pm$ 0.0019} & \textbf{0.7257 $\pm$ 0.0146} & \textbf{0.9025 $\pm$ 0.0096} \\
\bottomrule
\end{tabular}
}
\end{table}

\subsection{Missing Modality Experiments}
We study robustness to missing modalities through a set of controlled corruption experiments designed to test how well MedMIX tolerates incomplete inputs during training and inference. These evaluations approximate incomplete-modality conditions in practice, although they do not fully capture real-world structured missingness. The main text focuses on MIMIC-IV-MM, and the appendix reports the full MedMIX--AdaCoMed comparison on OpenI, MIMIC-IV-MM, and MMIST-ccRCC. We describe the specific corruption protocols in the subsections below.

\begin{figure}[h]
\centering
\includegraphics[width=\columnwidth]{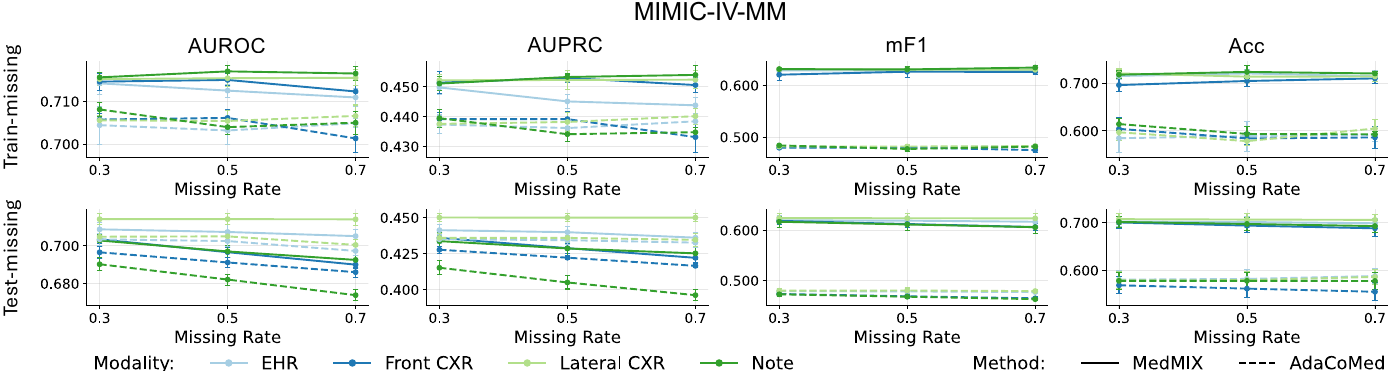}
\caption{MedMIX remains relatively stable under train-time one-modality drop and shows larger, but still controlled, degradation at test time, while consistently outperforming AdaCoMed. The top panels show train-time one-modality drop, where models are trained with one designated modality removed at rate $p\in\{0.3,0.5,0.7\}$ and evaluated on the standard test split; the bottom panels show test-time one-modality drop with the same per-modality rates. Subplots report AUROC, AUPRC, mF1, and Acc for frontal CXR, lateral CXR, EHR, and Note.}
\label{fig:train_test_onemod_mimic}
\end{figure}

\noindent\textbf{Single-modality drop under train-time and test-time corruption.}
\emph{Train missing.} We remove one modality at a time during training with rate $p\in\{0.3,0.5,0.7\}$ and evaluate on the standard test split. Figure~\ref{fig:train_test_onemod_mimic} summarizes the MIMIC-IV-MM one-modality-drop results for both train-time and test-time settings; Appendix Figures~\ref{fig:appendix_train_missing_onemod_openi} and \ref{fig:appendix_train_missing_onemod_mmist} and Appendix Table~\ref{tab:trainmissing_onemod_methods} report the corresponding OpenI and MMIST-ccRCC train-time results. On MIMIC-IV-MM, MedMIX remains stable, with AUROC ranging from 0.7109 to 0.7170 across dropped modalities and rates and mF1 from 0.6205 to 0.6343. Across datasets, MedMIX consistently outperforms AdaCoMed under this protocol.

\emph{Test missing.} We also report one-modality-at-a-time missingness at test time; the test-time panels in Figure~\ref{fig:train_test_onemod_mimic}, together with Appendix Figures~\ref{fig:appendix_test_missing_onemod_openi} and \ref{fig:appendix_test_missing_onemod_mmist} and Appendix Table~\ref{tab:testmissing_onemod_medmix_vs_adacomed}, provide the full comparison. Here the modality-specific dependencies become clearer. Modality-specific dependencies vary by dataset; full per-modality breakdowns are in the appendix.

Overall, the single-modality-drop results indicate that MedMIX is not uniformly insensitive to missing inputs, but it copes better with the loss of a single modality during training than at test time. Across datasets, removing clinical notes most often causes the largest performance drop, whereas some image views, such as the OpenI side-view CXR and MIMIC-IV-MM lateral CXR, appear relatively redundant under this ablation. Nevertheless, MedMIX consistently outperforms AdaCoMed across all appendix comparisons.

\noindent\textbf{Multi-random drop under train-time and test-time corruption.} This protocol targets robustness under Bernoulli-sampled missingness when the missing pattern is not known in advance.

\begin{figure}[h]
\centering
\includegraphics[width=\columnwidth]{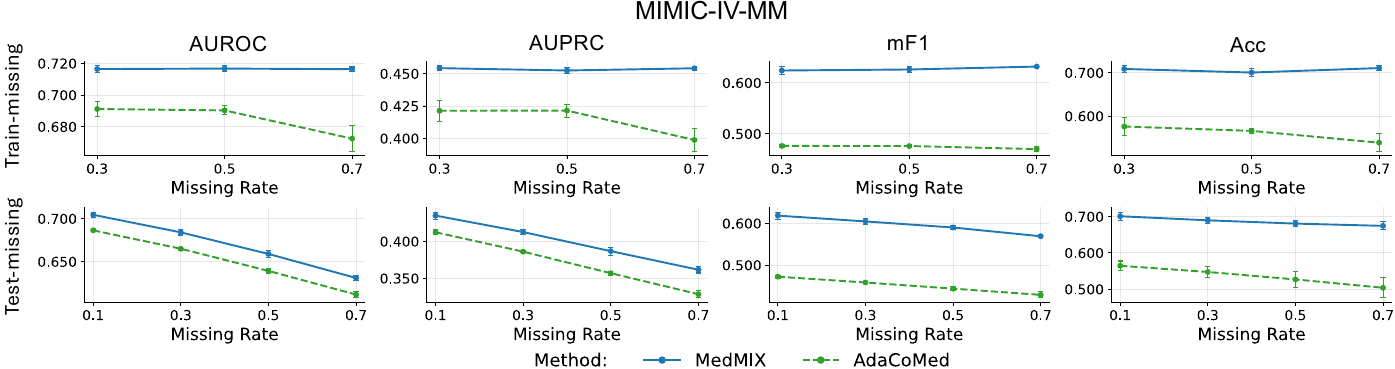}
\caption{MedMIX is robust to train-time multi-random drop and degrades gracefully as test-time missingness increases, while remaining stronger than AdaCoMed across all evaluated rates. The top panels show train-time multi-random drop, where each modality is independently dropped with probability $p\in\{0.1,0.3,0.5,0.7\}$ during training; the bottom panels show test-time multi-random drop with evaluation-time drop probability $r\in\{0.1,0.3,0.5,0.7\}$. Subplots report AUROC, AUPRC, mF1, and Acc.}
\label{fig:train_test_multi_mimic}
\end{figure}

\emph{Train missing.} During training, each modality is independently dropped with probability $p\in\{0.1,0.3,0.5,0.7\}$. Figure~\ref{fig:train_test_multi_mimic} summarizes the MIMIC-IV-MM multi-random-drop sweep for both train-time and test-time settings. Appendix Figures~\ref{fig:appendix_train_missing_multi_openi} and \ref{fig:appendix_train_missing_multi_mmist} show the corresponding train-time multi-random-drop sweeps on OpenI and MMIST-ccRCC, while Appendix Table~\ref{tab:trainmissing_multi_medmix_vs_adacomed} reports the full numeric comparison. MIMIC-IV-MM and OpenI remain highly stable across rates, while MMIST-ccRCC shows more visible degradation at higher $p$. MedMIX nevertheless remains ahead of AdaCoMed at every reported rate in all three datasets.

\emph{Test missing.} We further sweep test-time corruption rates $r\in\{0.1,0.3,0.5,0.7\}$. The test-time panels in Figure~\ref{fig:train_test_multi_mimic}, together with Appendix Figures~\ref{fig:appendix_test_missing_multi_openi} and \ref{fig:appendix_test_missing_multi_mmist} and Appendix Table~\ref{tab:testmissing_multi_random_methods}, report the full numeric comparison. They show a monotonic degradation with increasing drop rate: on MIMIC-IV-MM, AUROC and mF1 degrade monotonically; the same pattern appears on OpenI and MMIST-ccRCC. Even so, MedMIX remains ahead of AdaCoMed at every tested rate, indicating graceful degradation rather than a reversal in method ranking.

Taken together, the multi-random-drop results show that MedMIX degrades gracefully under stochastic missingness and maintains an advantage over AdaCoMed under the controlled protocols evaluated here, but they should not be interpreted as a comprehensive test of real-world structured missingness.


\subsection{Structural Ablation}
\begin{figure}[h]
\centering
\includegraphics[width=\textwidth]{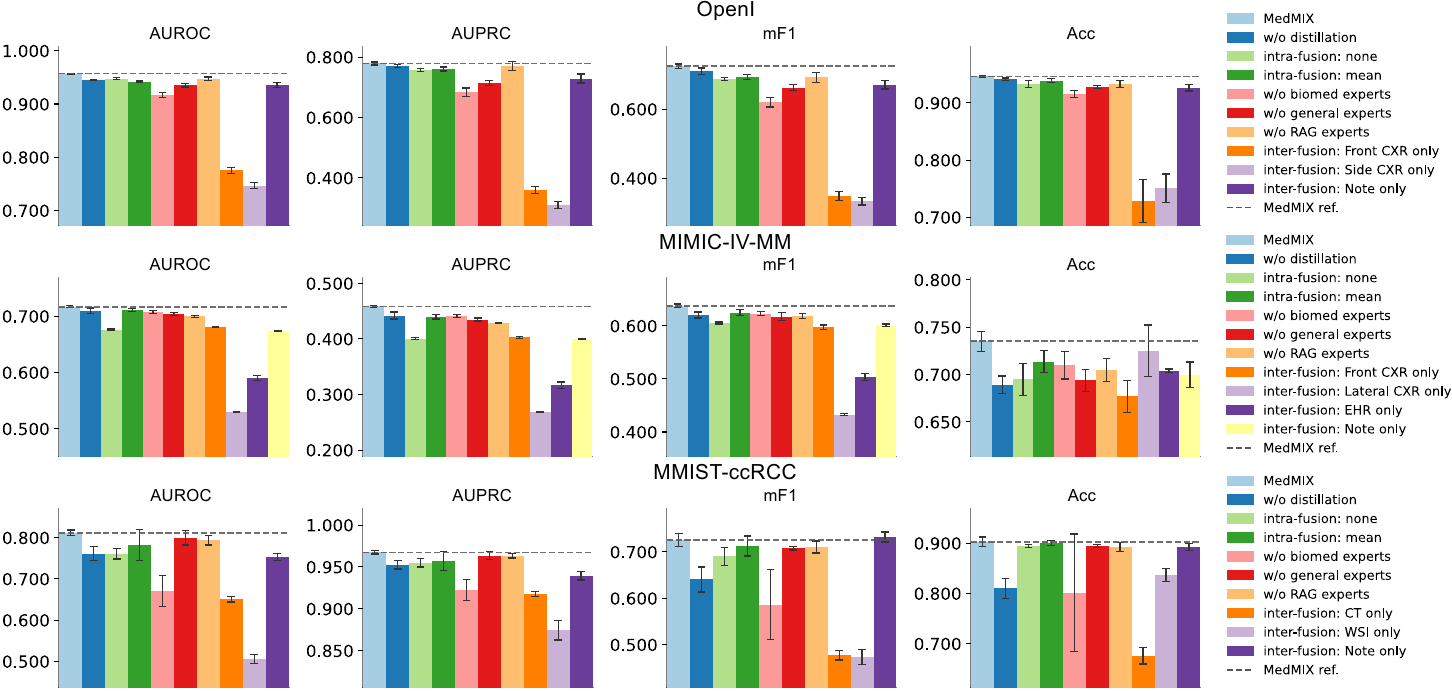}
\caption{Structural ablation results across OpenI, MIMIC-IV-MM, and MMIST-ccRCC.}
\label{fig:struct_ablation_main}
\end{figure}

To isolate the contribution of each architectural component, we ablate MedMIX on
OpenI, MIMIC-IV-MM, and MMIST-ccRCC. Figure~\ref{fig:struct_ablation_main}
visualizes the main trends. Appendix Table~\ref{tab:struct_ablation_three_datasets}
lists all numbers. We compare the full model against variants that (i) disable
distillation, (ii) drop one of the biomedical, general, or RAG SM expert
families within each modality, (iii) replace the intra-modality MoE with either
uniform averaging over the available SM experts (denoted \emph{uniform mean})
or the best SM expert alone,
and (iv) retain a single modality instead of inter-modality fusion.
MedMIX is best on 11 of the 12 dataset--metric pairs and second on the remaining
one. Removing distillation consistently degrades performance, with the largest drop
on MMIST-ccRCC. Both intra-modality substitutes lower AUROC, AUPRC, and mF1 on
every dataset, showing that the learned router is not replaceable by a fixed
rule or by any single SM expert. Dropping a single SM expert family is also harmful,
but the sensitivity is dataset-specific. OpenI and MMIST-ccRCC rely most on the
biomedical SM experts, while on MIMIC-IV-MM the loss is spread more evenly across
the three SM expert families. Restricting the model to any single modality produces much
larger drops. The Note-only variant marginally exceeds MedMIX on MMIST-ccRCC
mF1 but trails on AUROC, AUPRC, and accuracy, and no single-modality variant
dominates MedMIX across metrics. Distillation, a diverse set of within-modality
SM experts, learned expert fusion, and inter-modality fusion all contribute to
the full performance.

\subsection{External Validation}
To assess cross-cohort generalization, we train the model on MIMIC-IV-MM and directly evaluate it on MIMIC-III using the same test protocol. As shown in Table~\ref{tab:mimic3_external_compare_updated}, MedMIX achieves the best overall performance among all compared methods across all four metrics. Compared with the strongest competing baseline for each metric, MedMIX improves AUROC from 0.5423 to 0.5718, AUPRC from 0.1359 to 0.1508, mF1 from 0.4306 to 0.4560, and Acc from 0.6114 to 0.7371. Since this experiment evaluates a single source-to-target shift, we interpret it as evidence of improved cross-cohort robustness in this setting rather than as a comprehensive claim about domain generalization.
\begin{table}[h]
\centering
\caption{External-validation results on MIMIC-III.}
\label{tab:mimic3_external_compare_updated}
\resizebox{0.8\linewidth}{!}{%
\begin{tabular}{lcccc}
\toprule
Method & AUROC & AUPRC & mF1 & Acc \\
\midrule
MeanAvg & 0.5387 $\pm$ 0.0034 & 0.1324 $\pm$ 0.0019 & 0.3517 $\pm$ 0.0159 & 0.5058 $\pm$ 0.0348 \\
Concat & 0.5280 $\pm$ 0.0025 & 0.1287 $\pm$ 0.0018 & 0.3997 $\pm$ 0.0167 & 0.5948 $\pm$ 0.0148 \\
Max & 0.5371 $\pm$ 0.0102 & 0.1319 $\pm$ 0.0041 & 0.3800 $\pm$ 0.0487 & 0.5534 $\pm$ 0.0712 \\
Attention & \underline{0.5423 $\pm$ 0.0067} & 0.1306 $\pm$ 0.0019 & 0.3608 $\pm$ 0.0152 & 0.5212 $\pm$ 0.0177 \\
M3Care & 0.5343 $\pm$ 0.0249 & 0.1282 $\pm$ 0.0106 & \underline{0.4306 $\pm$ 0.0438} & 0.6093 $\pm$ 0.0706 \\
AdaCoMed & 0.5240 $\pm$ 0.0114 & \underline{0.1359 $\pm$ 0.0064} & 0.3916 $\pm$ 0.0243 & \underline{0.6114 $\pm$ 0.0433} \\
OneLLM & 0.5323 $\pm$ 0.0093 & 0.1291 $\pm$ 0.0056 & 0.3662 $\pm$ 0.0152 & 0.6073 $\pm$ 0.0566 \\
MedMIX & \textbf{0.5718 $\pm$ 0.0049} & \textbf{0.1508 $\pm$ 0.0034} & \textbf{0.4560 $\pm$ 0.0100} & \textbf{0.7371 $\pm$ 0.0257} \\
\bottomrule
\end{tabular}%
}
\end{table}

\subsection{Efficiency Analysis}
\begin{figure}[h]
\centering
\includegraphics[width=\textwidth]{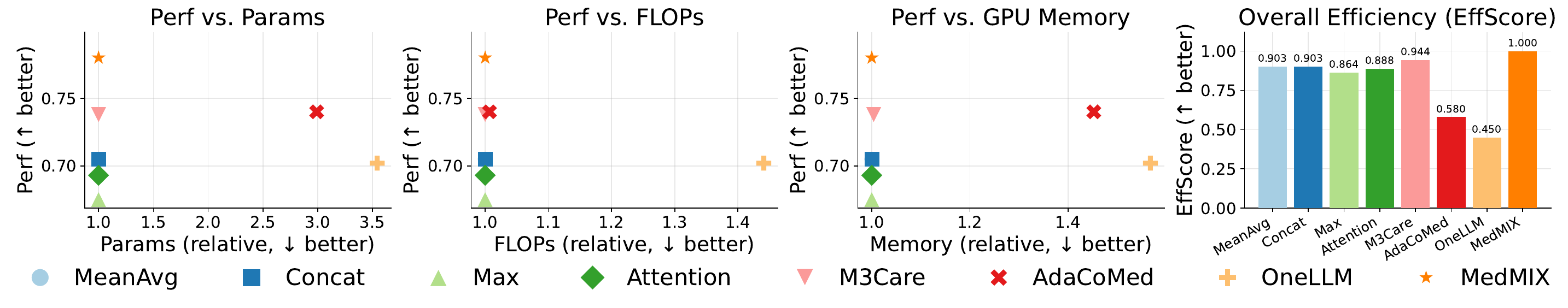}
\caption{Macro-averaged efficiency comparison across OpenI, MIMIC-IV-MM, and MMIST-ccRCC. The first three panels plot average predictive performance (Perf) against relative parameter count, FLOPs, and peak GPU memory. The fourth panel summarizes the resulting efficiency score (EffScore). Lower cost and higher performance are better.}
\label{fig:efficiency_tradeoff}
\end{figure}

We compare deployment-time parameters, inference FLOPs, and peak GPU memory across all methods. As shown in Figure~\ref{fig:efficiency_tradeoff} and Appendix Table~\ref{tab:efficiency_all}, MedMIX lies on the Pareto frontier in the performance--cost plane and achieves the highest \emph{EffScore}. This score normalizes average predictive performance by the geometric mean of parameter, FLOP, and memory cost ratios relative to MedMIX, with the full definition provided in~\ref{sec:appendix_results_tables}. MedMIX outperforms both comparable-cost baselines and higher-cost methods such as AdaCoMed and OneLLM, despite their substantially larger parameter counts and inference overhead.

\section{Conclusion}
In this paper, we propose MedMIX for robust multimodal medical diagnosis with incomplete clinical modalities, comprising three core components: modality-internal expert fusion, missing-aware inter-modality fusion, and training-only large--small model distillation. Experiments across diverse clinical benchmarks demonstrate its effectiveness, and ablation studies validate each component's contribution. Broader societal implications are discussed in \ref{sec:appendix_broader_impacts}.

However, our evaluation relies on controlled random missingness, which may differ in real-world settings. Future work will extend MedMIX to structured missingness patterns and broader deployment scenarios, exploring more adaptive expert selection to improve robustness and generalization.

%
%


\bibliographystyle{unsrtnat}
\bibliography{neurips_2026}

\appendix

\renewcommand{\thesection}{Appendix \Alph{section}}
\makeatletter
\@addtoreset{table}{section}
\@addtoreset{figure}{section}
\makeatother
\setcounter{table}{0}
\setcounter{figure}{0}
\renewcommand{\thetable}{\Alph{section}\arabic{table}}
\renewcommand{\thefigure}{\Alph{section}\arabic{figure}}
\renewcommand{\theHtable}{appendix.\Alph{section}.\arabic{table}}
\renewcommand{\theHfigure}{appendix.\Alph{section}.\arabic{figure}}

\makeatletter
\renewcommand{\fnum@table}{Appendix Table~\thetable}
\renewcommand{\fnum@figure}{Appendix Figure~\thefigure}
\makeatother

\section{CT Volume Preprocessing for 2D Expert Models}
\label{sec:appendix_ct}
CT volumes in MMIST-ccRCC present a challenge for 2D foundation models such as
DINOv2: they are pretrained on natural images and accept a single
$224{\times}224$ RGB frame, whereas CT data consists of volumetric arrays of
shape $[D, H, W]$ (axial slices first) or $[H, W, D]$ (depth last), stored as
\texttt{.npy}, \texttt{.npz}, or \texttt{.nii.gz} files.
We handle this mismatch with a \emph{2.5D slice aggregation} pipeline.
The Merlin Biomedical expert and the RadFM teacher operate natively on 3D
volumes and require no such adaptation.

Given a CT volume with $D$ axial slices, we uniformly sample $N{=}48$ indices
spanning the full depth range:
\begin{equation}
    \mathcal{I} = \Bigl\{\Bigl\lfloor \tfrac{D-1}{N-1}\,i \Bigr\rfloor\Bigr\}_{i=0}^{N-1}.
\end{equation}
Each sampled slice $V[\mathcal{I}_i] \in \mathbb{R}^{H \times W}$ is
intensity-clipped to the Hounsfield Unit range $[-1000, 1000]$ and linearly
rescaled to $[0, 1]$,
converted to a three-channel RGB image by replicating the grayscale channel,
and resized to $224{\times}224$ using bicubic interpolation.
Each slice image is then independently encoded by DINOv2-B to obtain a
$768$-dimensional \texttt{[CLS]} token embedding, and the $N$ per-slice
embeddings are mean-pooled to produce a single volume-level representation.
Uniform sampling is preferred over center-cropping because ccRCC tumors can
appear at any depth level and center-only slices may miss peripheral lesions.

\section{Retrieval-Augmented Expert Embedding Construction}
\label{sec:appendix_rag}

We describe the construction of the retrieval-augmented (RAG) expert embeddings.
All RAG embeddings are pre-computed offline and then cached as fixed features; no retrieval occurs at inference time, so the retrieval stage itself is excluded from deployment-time efficiency accounting.
The same cached RAG embeddings are supplied to every compared method as part of the shared expert set, so no baseline is disadvantaged by lacking the report-bridging, MedlinePlus retrieval, or Qwen2 encoding pipeline. Throughout the paper, this RAG branch is counted as one expert within the underlying modality, used to emulate same-modality model diversity rather than to define a new modality.
The pipeline follows a unified two-stage structure---\emph{report bridging} followed
by \emph{knowledge retrieval and encoding}---but the bridging model differs by
modality to match the underlying data type.

\paragraph{EHR serialization.}
EHR data consists of structured tabular records with heterogeneous clinical
variables (laboratory values, vital signs, demographic fields).
Since MedCPT operates on natural language, we first serialize each patient's
EHR row into a key--value text string of the form
\texttt{``variable\_name: value; variable\_name: value; \ldots''},
preserving all 46 clinical variables in a fixed field order.
The serialized string then enters Stage~2 directly, bypassing the image
report-bridging step. This allows EHR records to be treated uniformly
within the same retrieval pipeline as text modalities, without any
learned imputation or modality-specific encoder head.

\paragraph{Stage 1: Modality-specific report bridging (image modalities only).}
Raw image inputs cannot be directly compared against a text knowledge base.
We therefore first generate a structured natural-language description of each image
using a modality-matched report-generation FM: MAIRA-2~\citep{hyland2023maira}
for chest X-rays (frontal and lateral), Merlin~\citep{blankemeier2024merlin} for
CT volumes, and HistGen~\citep{histgen} for whole-slide pathology images.
Each FM is chosen because it was pretrained on large-scale paired
image--report corpora for its respective imaging modality, producing
clinically grounded descriptions that capture modality-specific morphological
detail. Text modalities (radiology notes, clinical notes) bypass this stage
entirely, as they are already in natural-language form.

\paragraph{Stage 2: Knowledge retrieval and encoding.}
The bridged or serialized text $t^{(m)}$ is encoded with
MedCPT~\citep{jin2023medcpt}, a contrastive retrieval encoder trained on
biomedical literature, to retrieve the top-$K$ most semantically similar
entries from the MedlinePlus health knowledge base.
The retrieved document embeddings are mean-pooled into a single retrieval context vector, then projected into Qwen2-0.5B's embedding space via a learned linear projection layer before being passed through the model as a soft prompt token, producing the final RAG embedding:
\begin{equation}
    e_{\mathrm{RAG}}^{(m)}
    = \mathrm{Qwen2\text{-}0.5B}\!\left(
        \frac{1}{K}\sum_{i=1}^{K}
        \mathrm{MedCPT}\bigl(\mathcal{N}_i(t^{(m)})\bigr)
      \right),
\end{equation}
where $\mathcal{N}_i(\cdot)$ denotes the $i$-th nearest neighbor in the
MedlinePlus index under MedCPT cosine similarity, with $K{=}5$ fixed across
all datasets and modalities. The full per-modality bridge model assignments
are summarized in Appendix Table~\ref{tab:experts}.

\paragraph{Data leakage considerations.}
No leakage arises from the RAG pipeline under our construction.
The MedlinePlus knowledge base is a static, publicly available consumer health
encyclopedia that contains no patient records, outcome labels, or split-specific
information; retrieval therefore cannot propagate any test-set signal into the
embeddings.
All RAG embeddings are pre-computed once from the raw inputs and
cached before any model training begins; the retrieval index and encoding
pipeline are never updated in response to training-set gradients or labels.
This ensures that the RAG expert embeddings are
strictly derived from observed input features and external static knowledge,
introducing no label or split leakage into any compared method.

\section{Reproducibility Details}\label{sec:appendix_reproducibility}
To facilitate reproduction, we summarize the key implementation choices and
evaluation settings used throughout the paper.

\paragraph{Dataset splits and modalities.}
OpenI uses a 0.65--0.15--0.20 train/val/test split with 15 binary radiographic
findings for multi-label classification with long-tailed prevalences.
MIMIC-IV-MM uses a 0.72--0.13--0.15 split, includes 46 EHR clinical variables,
and predicts three binary clinical outcomes: in-hospital mortality, long stay
$\geq$7\,d, and 30-day readmission, with prevalences of 16--35\%.
MMIST-ccRCC follows the official split and is formulated as a highly imbalanced
binary task, namely 12-month vital status with 88\% positive cases.
MIMIC-III is used only for external validation on the same three outcomes as
MIMIC-IV-MM.

\paragraph{Expert embeddings.}
For each modality we extract three complementary expert embeddings with small models (SM):
a general-purpose SM (\emph{General SM Expert}), a domain-specific biomedical SM
(\emph{Biomed SM Expert}), and a retrieval-augmented embedding (\emph{RAG SM Expert}).
The RAG expert is a fixed member of the shared expert committee for all compared methods,
serving as an additional same-modality expert branch rather than a separate modality.
Teacher LM embeddings are used for distillation only and are excluded at inference.
The full per-modality expert and teacher model assignments are listed in Appendix
Table~\ref{tab:experts}.

\paragraph{Optimization and training.}
All models are trained in PyTorch on NVIDIA A100 GPUs for up to 200 epochs
with mixed-precision. We use the AdamW optimizer (weight decay
$1\times10^{-2}$, gradient-norm clipping at $1.0$, $\beta_1{=}0.9$,
$\beta_2{=}0.999$). The base learning rate is $\eta_{\mathrm{base}}=1\times
10^{-5}$, and the router learning rate is set to $0.3\,\eta_{\mathrm{base}}$,
with a linear warm-up over the first $50$ epochs. The distillation weight
is linearly ramped from $0$ to $\lambda_D^{\max}=0.3$ over the first $T_D=30$
epochs and then held constant, while the relational knowledge distillation
term uses $\lambda_{\mathrm{RKD}}=0.05$ throughout training. We use a batch
size of $256$ across all three datasets. We select the checkpoint with the
best validation loss and stop early after $20$ epochs without
improvement. Results are reported over $5$ random seeds.

\paragraph{Baseline setup.}
All embedding-based baselines use the same expert definitions as MedMIX. In
particular, MeanAvg, Concat, Max, Attention, and M3Care all use the identical
per-modality expert set, including the same retrieval-augmented branch when
applicable. For the efficiency analysis, we count the deployed upstream expert
feature constructors together with the downstream fusion module; only
offline-only components, such as the cached retrieval stage and training-only
teacher encoders, are excluded. Thus, differences among these methods reflect
the fusion strategy rather than unequal access to preprocessing or external
knowledge.

\section{Broader Impacts}
\label{sec:appendix_broader_impacts}

\noindent\textbf{Societal Implications.}
Multimodal clinical AI systems are often developed and validated under idealized conditions of complete data availability, yet real-world healthcare settings are characterized by persistent modality incompleteness due to infrastructure constraints, cost barriers, and differential access to diagnostic procedures. This disparity means that models requiring complete modalities at inference time systematically underperform for the populations that stand to benefit most from automated clinical decision support. MedMIX directly addresses this gap by maintaining robust predictive performance under arbitrary missing-modality conditions, lowering the data-completeness requirements for deployment. Beyond data completeness, computational cost poses an equally significant barrier to adoption, and by retaining a lightweight inference pathway that excludes large teacher models at deployment, MedMIX lowers this barrier without sacrificing predictive performance. By decoupling model performance from the availability of expensive imaging or laboratory modalities, this work contributes toward more equitable clinical AI deployment across heterogeneous healthcare systems globally.

\noindent\textbf{Technical Generalizability.}
Beyond the specific clinical tasks studied here, MedMIX introduces a generalizable framework for orchestrating heterogeneous foundation model experts within and across modalities. As the landscape of domain-specific foundation models continues to expand, the challenge of principled multi-expert aggregation is not unique to medical AI: it arises in any domain where multiple pretrained specialists offer complementary views of the same input. The intra-modality MoE aggregation, masked inter-modality fusion, and training-only distillation components of MedMIX are modality-agnostic by design and can be adapted to other multimodal prediction settings, including scientific discovery, remote sensing, and embodied AI, where modality incompleteness and model ecosystem diversity are similarly prevalent.

\section{Additional Result Tables}\label{sec:appendix_results_tables}
\begin{table}[!h]
\centering
\caption{Expert and teacher foundation models by modality and dataset.}
\label{tab:experts}
\resizebox{\columnwidth}{!}{%
\begin{tabular}{llcccc}
\toprule
Dataset & Modality & Biomed & General & RAG & Teacher FM \\
\midrule
\multirow{3}{*}{OpenI}
 & Frontal CXR    & BiomedCLIP  & DINOv2-B   & MAIRA-2+MedCPT+Qwen2-0.5B & DINOv3 ViT-L    \\
 & Lateral CXR    & BiomedCLIP  & DINOv2-B   & MAIRA-2+MedCPT+Qwen2-0.5B & DINOv3 ViT-L    \\
 & Radiology Note & GatorTron-B & Qwen2-0.5B & MedCPT+Qwen2-0.5B         & Qwen2-7B  \\
\midrule
\multirow{4}{*}{MIMIC-IV-MM}
 & Frontal CXR    & BiomedCLIP  & DINOv2-B   & MAIRA-2+MedCPT+Qwen2-0.5B & DINOv3 ViT-L    \\
 & Lateral CXR    & BiomedCLIP  & DINOv2-B   & MAIRA-2+MedCPT+Qwen2-0.5B & DINOv3 ViT-L    \\
 & EHR            & MOIRAI      & Chronos-S  & MedCPT+Qwen2-0.5B         & Chronos-L \\
 & Radiology Note & GatorTron-B & Qwen2-0.5B & MedCPT+Qwen2-0.5B         & Qwen2-7B  \\
\midrule
\multirow{3}{*}{MMIST-ccRCC}
 & CT             & Merlin      & DINOv2-B   & Merlin+MedCPT+Qwen2-0.5B  & RadFM     \\
 & WSI            & BiomedCLIP  & DINOv2-B   & HistGen+MedCPT+Qwen2-0.5B & DINOv3 ViT-L    \\
 & Clinical Note  & GatorTron-B & Qwen2-0.5B & MedCPT+Qwen2-0.5B         & Qwen2-7B  \\
\midrule
\multirow{2}{*}{MIMIC-III}
 & EHR            & MOIRAI      & Chronos-S  & MedCPT+Qwen2-0.5B         & Chronos-L \\
 & Radiology Note & GatorTron-B & Qwen2-0.5B & MedCPT+Qwen2-0.5B         & Qwen2-7B  \\
\bottomrule
\end{tabular}%
}
\end{table}

\begin{table*}[!htbp]
\centering
\caption{Train-time one-modality-drop robustness: MedMIX vs.\ AdaCoMed across datasets and missing rates.}
\label{tab:trainmissing_onemod_methods}
\scriptsize
\setlength{\tabcolsep}{2.8pt}
\renewcommand{\arraystretch}{1.06}
\resizebox{\textwidth}{!}{
\begin{tabular}{llcccccc}
\toprule
\textbf{Dataset} & \textbf{Method} & \textbf{Missing Rate} & \textbf{Modality} & \textbf{AUROC} & \textbf{AUPRC} & \textbf{mF1} & \textbf{Acc} \\
\midrule
\multirow{18}{*}{OpenI}
& \multirow{9}{*}{MedMIX}
& \multirow{3}{*}{0.3} & Front CXR & 0.9458$\pm$0.0017 & 0.7739$\pm$0.0083 & 0.7121$\pm$0.0129 & 0.9412$\pm$0.0091 \\
& & & Side CXR  & 0.9465$\pm$0.0019 & 0.7770$\pm$0.0095 & 0.7194$\pm$0.0127 & 0.9453$\pm$0.0045 \\
& & & Note      & 0.9548$\pm$0.0005 & 0.7911$\pm$0.0038 & 0.7161$\pm$0.0056 & 0.9416$\pm$0.0016 \\
& & \multirow{3}{*}{0.5} & Front CXR & 0.9464$\pm$0.0018 & 0.7766$\pm$0.0091 & 0.7144$\pm$0.0156 & 0.9419$\pm$0.0098 \\
& & & Side CXR  & 0.9466$\pm$0.0018 & 0.7782$\pm$0.0067 & 0.7142$\pm$0.0063 & 0.9426$\pm$0.0051 \\
& & & Note      & 0.9542$\pm$0.0007 & 0.7870$\pm$0.0028 & 0.7138$\pm$0.0089 & 0.9404$\pm$0.0032 \\
& & \multirow{3}{*}{0.7} & Front CXR & 0.9457$\pm$0.0021 & 0.7738$\pm$0.0102 & 0.7045$\pm$0.0175 & 0.9358$\pm$0.0101 \\
& & & Side CXR  & 0.9461$\pm$0.0020 & 0.7742$\pm$0.0094 & 0.7165$\pm$0.0126 & 0.9442$\pm$0.0040 \\
& & & Note      & 0.9513$\pm$0.0015 & 0.7732$\pm$0.0035 & 0.6984$\pm$0.0016 & 0.9329$\pm$0.0028 \\
\cmidrule(lr){2-8}
& \multirow{9}{*}{AdaCoMed}
& \multirow{3}{*}{0.3} & Front CXR & 0.9063$\pm$0.0034 & 0.6012$\pm$0.0118 & 0.5659$\pm$0.0105 & 0.8944$\pm$0.0083 \\
& & & Side CXR  & 0.9110$\pm$0.0028 & 0.6114$\pm$0.0148 & 0.5734$\pm$0.0119 & 0.8958$\pm$0.0083 \\
& & & Note      & 0.9006$\pm$0.0076 & 0.5741$\pm$0.0242 & 0.5447$\pm$0.0207 & 0.8877$\pm$0.0106 \\
& & \multirow{3}{*}{0.5} & Front CXR & 0.9060$\pm$0.0044 & 0.6026$\pm$0.0155 & 0.5686$\pm$0.0145 & 0.8909$\pm$0.0085 \\
& & & Side CXR  & 0.9101$\pm$0.0035 & 0.6093$\pm$0.0160 & 0.5702$\pm$0.0169 & 0.8963$\pm$0.0089 \\
& & & Note      & 0.8885$\pm$0.0077 & 0.5319$\pm$0.0245 & 0.5153$\pm$0.0221 & 0.8778$\pm$0.0064 \\
& & \multirow{3}{*}{0.7} & Front CXR & 0.9061$\pm$0.0028 & 0.6058$\pm$0.0132 & 0.5624$\pm$0.0145 & 0.8862$\pm$0.0111 \\
& & & Side CXR  & 0.9096$\pm$0.0033 & 0.6081$\pm$0.0156 & 0.5632$\pm$0.0150 & 0.8865$\pm$0.0063 \\
& & & Note      & 0.8666$\pm$0.0081 & 0.4719$\pm$0.0257 & 0.4558$\pm$0.0234 & 0.8448$\pm$0.0131 \\
\midrule
\multirow{24}{*}{MIMIC-IV-MM}
& \multirow{12}{*}{MedMIX}
& \multirow{4}{*}{0.3} & Front CXR   & 0.7146$\pm$0.0020 & 0.4514$\pm$0.0037 & 0.6205$\pm$0.0112 & 0.6958$\pm$0.0133 \\
& & & Lateral CXR & 0.7152$\pm$0.0016 & 0.4522$\pm$0.0020 & 0.6313$\pm$0.0027 & 0.7201$\pm$0.0092 \\
& & & EHR         & 0.7142$\pm$0.0026 & 0.4498$\pm$0.0025 & 0.6286$\pm$0.0023 & 0.7147$\pm$0.0073 \\
& & & Note        & 0.7156$\pm$0.0011 & 0.4511$\pm$0.0023 & 0.6313$\pm$0.0033 & 0.7175$\pm$0.0077 \\
& & \multirow{4}{*}{0.5} & Front CXR   & 0.7150$\pm$0.0014 & 0.4530$\pm$0.0016 & 0.6263$\pm$0.0105 & 0.7043$\pm$0.0117 \\
& & & Lateral CXR & 0.7154$\pm$0.0019 & 0.4523$\pm$0.0032 & 0.6303$\pm$0.0040 & 0.7134$\pm$0.0113 \\
& & & EHR         & 0.7125$\pm$0.0015 & 0.4451$\pm$0.0023 & 0.6285$\pm$0.0029 & 0.7201$\pm$0.0087 \\
& & & Note        & 0.7170$\pm$0.0014 & 0.4533$\pm$0.0010 & 0.6307$\pm$0.0063 & 0.7233$\pm$0.0136 \\
& & \multirow{4}{*}{0.7} & Front CXR   & 0.7123$\pm$0.0014 & 0.4506$\pm$0.0024 & 0.6259$\pm$0.0052 & 0.7097$\pm$0.0101 \\
& & & Lateral CXR & 0.7155$\pm$0.0019 & 0.4524$\pm$0.0030 & 0.6301$\pm$0.0039 & 0.7137$\pm$0.0105 \\
& & & EHR         & 0.7109$\pm$0.0021 & 0.4438$\pm$0.0027 & 0.6281$\pm$0.0022 & 0.7141$\pm$0.0113 \\
& & & Note        & 0.7165$\pm$0.0016 & 0.4540$\pm$0.0032 & 0.6343$\pm$0.0042 & 0.7200$\pm$0.0044 \\
\cmidrule(lr){2-8}
& \multirow{12}{*}{AdaCoMed}
& \multirow{4}{*}{0.3} & Front CXR   & 0.7057$\pm$0.0014 & 0.4392$\pm$0.0022 & 0.4794$\pm$0.0019 & 0.6043$\pm$0.0238 \\
& & & Lateral CXR & 0.7056$\pm$0.0012 & 0.4376$\pm$0.0013 & 0.4817$\pm$0.0020 & 0.5971$\pm$0.0171 \\
& & & EHR         & 0.7044$\pm$0.0044 & 0.4374$\pm$0.0032 & 0.4799$\pm$0.0035 & 0.5848$\pm$0.0302 \\
& & & Note        & 0.7081$\pm$0.0016 & 0.4395$\pm$0.0031 & 0.4839$\pm$0.0025 & 0.6143$\pm$0.0125 \\
& & \multirow{4}{*}{0.5} & Front CXR   & 0.7061$\pm$0.0020 & 0.4392$\pm$0.0027 & 0.4798$\pm$0.0041 & 0.5844$\pm$0.0123 \\
& & & Lateral CXR & 0.7054$\pm$0.0024 & 0.4382$\pm$0.0030 & 0.4817$\pm$0.0035 & 0.5787$\pm$0.0086 \\
& & & EHR         & 0.7032$\pm$0.0032 & 0.4362$\pm$0.0018 & 0.4798$\pm$0.0030 & 0.5891$\pm$0.0313 \\
& & & Note        & 0.7040$\pm$0.0016 & 0.4341$\pm$0.0025 & 0.4770$\pm$0.0054 & 0.5937$\pm$0.0159 \\
& & \multirow{4}{*}{0.7} & Front CXR   & 0.7013$\pm$0.0033 & 0.4331$\pm$0.0052 & 0.4748$\pm$0.0049 & 0.5864$\pm$0.0231 \\
& & & Lateral CXR & 0.7066$\pm$0.0026 & 0.4401$\pm$0.0027 & 0.4825$\pm$0.0016 & 0.6051$\pm$0.0188 \\
& & & EHR         & 0.7049$\pm$0.0011 & 0.4384$\pm$0.0012 & 0.4825$\pm$0.0020 & 0.5915$\pm$0.0137 \\
& & & Note        & 0.7050$\pm$0.0027 & 0.4348$\pm$0.0017 & 0.4816$\pm$0.0035 & 0.5929$\pm$0.0068 \\
\midrule
\multirow{18}{*}{MMIST-ccRCC}
& \multirow{9}{*}{MedMIX}
& \multirow{3}{*}{0.3} & CT   & 0.7930$\pm$0.0104 & 0.9622$\pm$0.0029 & 0.7278$\pm$0.0065 & 0.9041$\pm$0.0041 \\
& & & WSI  & 0.7941$\pm$0.0139 & 0.9628$\pm$0.0033 & 0.7067$\pm$0.0194 & 0.8959$\pm$0.0112 \\
& & & Note & 0.7778$\pm$0.0118 & 0.9594$\pm$0.0026 & 0.7055$\pm$0.0089 & 0.8926$\pm$0.0000 \\
& & \multirow{3}{*}{0.5} & CT   & 0.7608$\pm$0.0152 & 0.9520$\pm$0.0056 & 0.7291$\pm$0.0059 & 0.9025$\pm$0.0062 \\
& & & WSI  & 0.7993$\pm$0.0154 & 0.9640$\pm$0.0035 & 0.7082$\pm$0.0136 & 0.8942$\pm$0.0062 \\
& & & Note & 0.7604$\pm$0.0135 & 0.9561$\pm$0.0031 & 0.6854$\pm$0.0082 & 0.8843$\pm$0.0091 \\
& & \multirow{3}{*}{0.7} & CT   & 0.7531$\pm$0.0084 & 0.9465$\pm$0.0051 & 0.7035$\pm$0.0425 & 0.8909$\pm$0.0160 \\
& & & WSI  & 0.7972$\pm$0.0175 & 0.9641$\pm$0.0038 & 0.7137$\pm$0.0217 & 0.8942$\pm$0.0151 \\
& & & Note & 0.7350$\pm$0.0130 & 0.9473$\pm$0.0061 & 0.6383$\pm$0.0749 & 0.8215$\pm$0.1341 \\
\cmidrule(lr){2-8}
& \multirow{9}{*}{AdaCoMed}
& \multirow{3}{*}{0.3} & CT   & 0.7361$\pm$0.0201 & 0.9434$\pm$0.0052 & 0.5578$\pm$0.0945 & 0.7554$\pm$0.1651 \\
& & & WSI  & 0.7170$\pm$0.0211 & 0.9374$\pm$0.0067 & 0.5563$\pm$0.0639 & 0.6810$\pm$0.1223 \\
& & & Note & 0.7334$\pm$0.0129 & 0.9436$\pm$0.0021 & 0.6075$\pm$0.0734 & 0.8595$\pm$0.0395 \\
& & \multirow{3}{*}{0.5} & CT   & 0.7283$\pm$0.0206 & 0.9409$\pm$0.0053 & 0.5903$\pm$0.0893 & 0.8231$\pm$0.1271 \\
& & & WSI  & 0.7031$\pm$0.0252 & 0.9317$\pm$0.0092 & 0.5425$\pm$0.0796 & 0.7372$\pm$0.1526 \\
& & & Note & 0.7308$\pm$0.0169 & 0.9422$\pm$0.0036 & 0.5584$\pm$0.0847 & 0.7554$\pm$0.1490 \\
& & \multirow{3}{*}{0.7} & CT   & 0.7266$\pm$0.0207 & 0.9394$\pm$0.0058 & 0.5461$\pm$0.0650 & 0.6810$\pm$0.1493 \\
& & & WSI  & 0.7042$\pm$0.0228 & 0.9322$\pm$0.0108 & 0.4806$\pm$0.0132 & 0.6810$\pm$0.1526 \\
& & & Note & 0.7110$\pm$0.0167 & 0.9317$\pm$0.0109 & 0.5201$\pm$0.0509 & 0.6215$\pm$0.1198 \\
\bottomrule
\end{tabular}
}
\end{table*}

\begin{table*}[!htbp]
\centering
\caption{Test-time one-modality-drop robustness: MedMIX vs.\ AdaCoMed across datasets and missing rates.}
\label{tab:testmissing_onemod_medmix_vs_adacomed}
\scriptsize
\setlength{\tabcolsep}{2.8pt}
\renewcommand{\arraystretch}{1.06}
\resizebox{\textwidth}{!}{
\begin{tabular}{llcccccc}
\toprule
\textbf{Dataset} & \textbf{Method} & \textbf{Missing Rate} & \textbf{Modality} & \textbf{AUROC} & \textbf{AUPRC} & \textbf{mF1} & \textbf{Acc} \\
\midrule
\multirow{18}{*}{OpenI}
& \multirow{9}{*}{MedMIX}
& \multirow{3}{*}{0.3} & Front CXR   & 0.9454$\pm$0.0019 & 0.7742$\pm$0.0080 & 0.7150$\pm$0.0113 & 0.9430$\pm$0.0049 \\
& & & Side CXR    & 0.9450$\pm$0.0018 & 0.7726$\pm$0.0075 & 0.7150$\pm$0.0113 & 0.9430$\pm$0.0049 \\
& & & Note        & 0.8728$\pm$0.0104 & 0.6307$\pm$0.0187 & 0.5814$\pm$0.0113 & 0.9214$\pm$0.0029 \\
& & \multirow{3}{*}{0.5} & Front CXR & 0.9453$\pm$0.0019 & 0.7738$\pm$0.0083 & 0.7150$\pm$0.0113 & 0.9430$\pm$0.0049 \\
& & & Side CXR    & 0.9452$\pm$0.0019 & 0.7735$\pm$0.0079 & 0.7150$\pm$0.0113 & 0.9430$\pm$0.0049 \\
& & & Note        & 0.8434$\pm$0.0113 & 0.5699$\pm$0.0243 & 0.5225$\pm$0.0146 & 0.9140$\pm$0.0027 \\
& & \multirow{3}{*}{0.7} & Front CXR & 0.9453$\pm$0.0016 & 0.7738$\pm$0.0080 & 0.7150$\pm$0.0113 & 0.9430$\pm$0.0049 \\
& & & Side CXR    & 0.9454$\pm$0.0018 & 0.7740$\pm$0.0077 & 0.7149$\pm$0.0113 & 0.9430$\pm$0.0049 \\
& & & Note        & 0.7949$\pm$0.0137 & 0.4952$\pm$0.0263 & 0.4386$\pm$0.0213 & 0.9050$\pm$0.0021 \\
\cmidrule(lr){2-8}
& \multirow{9}{*}{AdaCoMed}
& \multirow{3}{*}{0.3} & Front CXR   & 0.9096$\pm$0.0052 & 0.6161$\pm$0.0206 & 0.5782$\pm$0.0175 & 0.8989$\pm$0.0088 \\
& & & Side CXR    & 0.9106$\pm$0.0051 & 0.6172$\pm$0.0209 & 0.5805$\pm$0.0175 & 0.8991$\pm$0.0088 \\
& & & Note        & 0.8624$\pm$0.0078 & 0.5310$\pm$0.0227 & 0.5005$\pm$0.0215 & 0.8824$\pm$0.0089 \\
& & \multirow{3}{*}{0.5} & Front CXR & 0.9087$\pm$0.0057 & 0.6163$\pm$0.0202 & 0.5784$\pm$0.0181 & 0.8997$\pm$0.0083 \\
& & & Side CXR    & 0.9112$\pm$0.0054 & 0.6189$\pm$0.0216 & 0.5816$\pm$0.0181 & 0.8996$\pm$0.0089 \\
& & & Note        & 0.8229$\pm$0.0059 & 0.4651$\pm$0.0216 & 0.4396$\pm$0.0154 & 0.8706$\pm$0.0077 \\
& & \multirow{3}{*}{0.7} & Front CXR & 0.9084$\pm$0.0063 & 0.6149$\pm$0.0208 & 0.5773$\pm$0.0187 & 0.8999$\pm$0.0080 \\
& & & Side CXR    & 0.9106$\pm$0.0056 & 0.6174$\pm$0.0216 & 0.5802$\pm$0.0193 & 0.8991$\pm$0.0098 \\
& & & Note        & 0.7790$\pm$0.0062 & 0.3960$\pm$0.0161 & 0.3733$\pm$0.0130 & 0.8590$\pm$0.0060 \\
\midrule
\multirow{24}{*}{MIMIC-IV-MM}
& \multirow{12}{*}{MedMIX}
& \multirow{4}{*}{0.3} & EHR         & 0.7087$\pm$0.0037 & 0.4413$\pm$0.0039 & 0.6198$\pm$0.0090 & 0.7022$\pm$0.0111 \\
& & & Front CXR   & 0.7034$\pm$0.0022 & 0.4360$\pm$0.0039 & 0.6187$\pm$0.0066 & 0.7003$\pm$0.0122 \\
& & & Lateral CXR & 0.7141$\pm$0.0029 & 0.4501$\pm$0.0028 & 0.6242$\pm$0.0092 & 0.7072$\pm$0.0119 \\
& & & Note        & 0.7028$\pm$0.0031 & 0.4337$\pm$0.0063 & 0.6169$\pm$0.0105 & 0.7014$\pm$0.0120 \\
& & \multirow{4}{*}{0.5} & EHR       & 0.7073$\pm$0.0042 & 0.4399$\pm$0.0042 & 0.6191$\pm$0.0091 & 0.7012$\pm$0.0111 \\
& & & Front CXR   & 0.6966$\pm$0.0026 & 0.4288$\pm$0.0034 & 0.6124$\pm$0.0062 & 0.6937$\pm$0.0137 \\
& & & Lateral CXR & 0.7141$\pm$0.0029 & 0.4500$\pm$0.0027 & 0.6239$\pm$0.0093 & 0.7063$\pm$0.0119 \\
& & & Note        & 0.6970$\pm$0.0031 & 0.4286$\pm$0.0049 & 0.6115$\pm$0.0115 & 0.6971$\pm$0.0128 \\
& & \multirow{4}{*}{0.7} & EHR       & 0.7051$\pm$0.0037 & 0.4361$\pm$0.0035 & 0.6169$\pm$0.0083 & 0.6986$\pm$0.0104 \\
& & & Front CXR   & 0.6900$\pm$0.0022 & 0.4221$\pm$0.0035 & 0.6064$\pm$0.0078 & 0.6880$\pm$0.0167 \\
& & & Lateral CXR & 0.7140$\pm$0.0031 & 0.4500$\pm$0.0027 & 0.6237$\pm$0.0092 & 0.7057$\pm$0.0118 \\
& & & Note        & 0.6925$\pm$0.0024 & 0.4252$\pm$0.0042 & 0.6061$\pm$0.0121 & 0.6927$\pm$0.0142 \\
\cmidrule(lr){2-8}
& \multirow{12}{*}{AdaCoMed}
& \multirow{4}{*}{0.3} & EHR         & 0.7034$\pm$0.0031 & 0.4350$\pm$0.0031 & 0.4789$\pm$0.0051 & 0.5805$\pm$0.0190 \\
& & & Front CXR   & 0.6965$\pm$0.0031 & 0.4277$\pm$0.0024 & 0.4732$\pm$0.0045 & 0.5687$\pm$0.0182 \\
& & & Lateral CXR & 0.7048$\pm$0.0032 & 0.4360$\pm$0.0032 & 0.4797$\pm$0.0052 & 0.5788$\pm$0.0177 \\
& & & Note        & 0.6902$\pm$0.0035 & 0.4152$\pm$0.0050 & 0.4726$\pm$0.0033 & 0.5780$\pm$0.0181 \\
& & \multirow{4}{*}{0.5} & EHR       & 0.7025$\pm$0.0031 & 0.4343$\pm$0.0033 & 0.4783$\pm$0.0052 & 0.5820$\pm$0.0195 \\
& & & Front CXR   & 0.6912$\pm$0.0026 & 0.4222$\pm$0.0014 & 0.4686$\pm$0.0045 & 0.5617$\pm$0.0183 \\
& & & Lateral CXR & 0.7050$\pm$0.0032 & 0.4360$\pm$0.0034 & 0.4801$\pm$0.0051 & 0.5795$\pm$0.0171 \\
& & & Note        & 0.6821$\pm$0.0029 & 0.4050$\pm$0.0045 & 0.4678$\pm$0.0025 & 0.5780$\pm$0.0179 \\
& & \multirow{4}{*}{0.7} & EHR       & 0.6973$\pm$0.0026 & 0.4327$\pm$0.0034 & 0.4770$\pm$0.0027 & 0.5880$\pm$0.0151 \\
& & & Front CXR   & 0.6860$\pm$0.0026 & 0.4165$\pm$0.0014 & 0.4646$\pm$0.0037 & 0.5551$\pm$0.0188 \\
& & & Lateral CXR & 0.7004$\pm$0.0031 & 0.4346$\pm$0.0045 & 0.4790$\pm$0.0024 & 0.5862$\pm$0.0150 \\
& & & Note        & 0.6738$\pm$0.0029 & 0.3962$\pm$0.0039 & 0.4629$\pm$0.0031 & 0.5777$\pm$0.0180 \\
\midrule
\multirow{18}{*}{MMIST-ccRCC}
& \multirow{9}{*}{MedMIX}
& \multirow{3}{*}{0.3} & CT          & 0.7880$\pm$0.0168 & 0.9616$\pm$0.0034 & 0.7267$\pm$0.0322 & 0.8959$\pm$0.0134 \\
& & & WSI         & 0.8036$\pm$0.0077 & 0.9642$\pm$0.0018 & 0.7280$\pm$0.0104 & 0.9041$\pm$0.0066 \\
& & & Note        & 0.7666$\pm$0.0517 & 0.9538$\pm$0.0185 & 0.6584$\pm$0.0290 & 0.8909$\pm$0.0081 \\
& & \multirow{3}{*}{0.5} & CT        & 0.7730$\pm$0.0237 & 0.9572$\pm$0.0052 & 0.7276$\pm$0.0372 & 0.8942$\pm$0.0142 \\
& & & WSI         & 0.7977$\pm$0.0021 & 0.9631$\pm$0.0004 & 0.7280$\pm$0.0104 & 0.9041$\pm$0.0066 \\
& & & Note        & 0.7500$\pm$0.0508 & 0.9509$\pm$0.0168 & 0.6193$\pm$0.0378 & 0.7702$\pm$0.0635 \\
& & \multirow{3}{*}{0.7} & CT        & 0.7605$\pm$0.0266 & 0.9537$\pm$0.0064 & 0.7083$\pm$0.0305 & 0.8843$\pm$0.0074 \\
& & & WSI         & 0.8012$\pm$0.0090 & 0.9639$\pm$0.0021 & 0.7253$\pm$0.0097 & 0.9025$\pm$0.0062 \\
& & & Note        & 0.7248$\pm$0.0494 & 0.9476$\pm$0.0146 & 0.5827$\pm$0.0183 & 0.8611$\pm$0.0269 \\
\cmidrule(lr){2-8}
& \multirow{9}{*}{AdaCoMed}
& \multirow{3}{*}{0.3} & CT          & 0.6983$\pm$0.0622 & 0.9309$\pm$0.0115 & 0.4906$\pm$0.1197 & 0.5950$\pm$0.2122 \\
& & & WSI         & 0.7162$\pm$0.0297 & 0.9381$\pm$0.0083 & 0.5680$\pm$0.0722 & 0.7058$\pm$0.1423 \\
& & & Note        & 0.7009$\pm$0.0297 & 0.9344$\pm$0.0081 & 0.5483$\pm$0.0689 & 0.6760$\pm$0.1270 \\
& & \multirow{3}{*}{0.5} & CT        & 0.7060$\pm$0.0692 & 0.9314$\pm$0.0180 & 0.4542$\pm$0.1529 & 0.5421$\pm$0.2515 \\
& & & WSI         & 0.7080$\pm$0.0119 & 0.9349$\pm$0.0062 & 0.5517$\pm$0.0516 & 0.7173$\pm$0.1393 \\
& & & Note        & 0.6966$\pm$0.0206 & 0.9318$\pm$0.0085 & 0.5361$\pm$0.0475 & 0.6793$\pm$0.1275 \\
& & \multirow{3}{*}{0.7} & CT        & 0.6965$\pm$0.0811 & 0.9295$\pm$0.0225 & 0.4113$\pm$0.1769 & 0.4942$\pm$0.2787 \\
& & & WSI         & 0.7046$\pm$0.0258 & 0.9349$\pm$0.0085 & 0.5424$\pm$0.0520 & 0.7289$\pm$0.1442 \\
& & & Note        & 0.6930$\pm$0.0215 & 0.9296$\pm$0.0122 & 0.5246$\pm$0.0422 & 0.6826$\pm$0.1372 \\
\bottomrule
\end{tabular}}
\end{table*}

\begin{table*}[!htbp]
\centering
\caption{Train-time multi-random-drop robustness: MedMIX vs.\ AdaCoMed across datasets and missing rates.}
\label{tab:trainmissing_multi_medmix_vs_adacomed}
\scriptsize
\setlength{\tabcolsep}{2.8pt}
\renewcommand{\arraystretch}{1.06}
\resizebox{\textwidth}{!}{
\begin{tabular}{llccccc}
\toprule
\textbf{Dataset} & \textbf{Method} & \textbf{Missing Rate} & \textbf{AUROC} & \textbf{AUPRC} & \textbf{mF1} & \textbf{Acc} \\
\midrule
\multirow{6}{*}{OpenI}
& \multirow{3}{*}{MedMIX}
& 0.3 & 0.9540$\pm$0.0012 & 0.7861$\pm$0.0074 & 0.7088$\pm$0.0128 & 0.9378$\pm$0.0040 \\
& & 0.5 & 0.9543$\pm$0.0003 & 0.7873$\pm$0.0029 & 0.7126$\pm$0.0029 & 0.9377$\pm$0.0028 \\
& & 0.7 & 0.9530$\pm$0.0015 & 0.7823$\pm$0.0080 & 0.6957$\pm$0.0085 & 0.9344$\pm$0.0042 \\
\cmidrule(lr){2-7}
& \multirow{3}{*}{AdaCoMed}
& 0.3 & 0.8559$\pm$0.0124 & 0.4596$\pm$0.0324 & 0.4246$\pm$0.0325 & 0.8176$\pm$0.0351 \\
& & 0.5 & 0.8402$\pm$0.0152 & 0.4254$\pm$0.0295 & 0.3919$\pm$0.0295 & 0.7473$\pm$0.0602 \\
& & 0.7 & 0.8163$\pm$0.0074 & 0.3748$\pm$0.0127 & 0.3652$\pm$0.0066 & 0.7243$\pm$0.0269 \\
\midrule
\multirow{6}{*}{MIMIC-IV-MM}
& \multirow{3}{*}{MedMIX}
& 0.3 & 0.7164$\pm$0.0020 & 0.4542$\pm$0.0020 & 0.6241$\pm$0.0070 & 0.7087$\pm$0.0074 \\
& & 0.5 & 0.7167$\pm$0.0019 & 0.4524$\pm$0.0026 & 0.6261$\pm$0.0050 & 0.7004$\pm$0.0094 \\
& & 0.7 & 0.7163$\pm$0.0014 & 0.4541$\pm$0.0017 & 0.6318$\pm$0.0020 & 0.7107$\pm$0.0058 \\
\cmidrule(lr){2-7}
& \multirow{3}{*}{AdaCoMed}
& 0.3 & 0.6912$\pm$0.0047 & 0.4213$\pm$0.0080 & 0.4762$\pm$0.0035 & 0.5758$\pm$0.0209 \\
& & 0.5 & 0.6903$\pm$0.0028 & 0.4215$\pm$0.0050 & 0.4760$\pm$0.0026 & 0.5658$\pm$0.0055 \\
& & 0.7 & 0.6725$\pm$0.0081 & 0.3988$\pm$0.0089 & 0.4700$\pm$0.0046 & 0.5385$\pm$0.0202 \\
\midrule
\multirow{6}{*}{MMIST-ccRCC}
& \multirow{3}{*}{MedMIX}
& 0.3 & 0.7512$\pm$0.0137 & 0.9515$\pm$0.0034 & 0.7020$\pm$0.0113 & 0.8992$\pm$0.0033 \\
& & 0.5 & 0.7373$\pm$0.0130 & 0.9458$\pm$0.0046 & 0.6619$\pm$0.0871 & 0.8281$\pm$0.1372 \\
& & 0.7 & 0.7314$\pm$0.0148 & 0.9436$\pm$0.0062 & 0.6548$\pm$0.0842 & 0.8231$\pm$0.1349 \\
\cmidrule(lr){2-7}
& \multirow{3}{*}{AdaCoMed}
& 0.3 & 0.7052$\pm$0.0278 & 0.9388$\pm$0.0056 & 0.4915$\pm$0.0195 & 0.6810$\pm$0.1559 \\
& & 0.5 & 0.6925$\pm$0.0237 & 0.9361$\pm$0.0015 & 0.4855$\pm$0.0107 & 0.6182$\pm$0.1249 \\
& & 0.7 & 0.7046$\pm$0.0245 & 0.9329$\pm$0.0116 & 0.5409$\pm$0.0493 & 0.6645$\pm$0.1214 \\
\bottomrule
\end{tabular}}
\end{table*}

\begin{table*}[!htbp]
\centering
\caption{Test-time multi-random-drop robustness: MedMIX vs.\ AdaCoMed across datasets and missing rates (mean$\pm$std over 5 seeds).}
\label{tab:testmissing_multi_random_methods}
\scriptsize
\setlength{\tabcolsep}{2.8pt}
\renewcommand{\arraystretch}{1.06}
\resizebox{\textwidth}{!}{
\begin{tabular}{llccccc}
\toprule
\textbf{Dataset} & \textbf{Method} & \textbf{Missing Rate} & \textbf{AUROC} & \textbf{AUPRC} & \textbf{mF1} & \textbf{Acc} \\
\midrule
\multirow{8}{*}{OpenI}
& \multirow{4}{*}{MedMIX}
 & 0.1 & 0.9239$\pm$0.0029 & 0.7246$\pm$0.0062 & 0.6710$\pm$0.0111 & 0.9353$\pm$0.0053 \\
&  & 0.3 & 0.8992$\pm$0.0039 & 0.6778$\pm$0.0080 & 0.6262$\pm$0.0113 & 0.9279$\pm$0.0043 \\
&  & 0.5 & 0.8708$\pm$0.0099 & 0.6287$\pm$0.0147 & 0.5810$\pm$0.0177 & 0.9196$\pm$0.0078 \\
&  & 0.7 & 0.8341$\pm$0.0097 & 0.5695$\pm$0.0114 & 0.5274$\pm$0.0144 & 0.9118$\pm$0.0097 \\
\cmidrule(lr){2-7}
& \multirow{4}{*}{AdaCoMed}
 & 0.1 & 0.8825$\pm$0.0098 & 0.5504$\pm$0.0236 & 0.5192$\pm$0.0284 & 0.8823$\pm$0.0087 \\
&  & 0.3 & 0.8460$\pm$0.0044 & 0.4974$\pm$0.0156 & 0.4749$\pm$0.0183 & 0.8778$\pm$0.0048 \\
&  & 0.5 & 0.8046$\pm$0.0106 & 0.4389$\pm$0.0265 & 0.4130$\pm$0.0351 & 0.8690$\pm$0.0050 \\
&  & 0.7 & 0.7772$\pm$0.0053 & 0.3912$\pm$0.0147 & 0.3699$\pm$0.0204 & 0.8626$\pm$0.0078 \\
\midrule
\multirow{8}{*}{MIMIC-IV-MM}
& \multirow{4}{*}{MedMIX}
 & 0.1 & 0.7045$\pm$0.0032 & 0.4349$\pm$0.0048 & 0.6185$\pm$0.0085 & 0.7008$\pm$0.0115 \\
&  & 0.3 & 0.6840$\pm$0.0031 & 0.4126$\pm$0.0036 & 0.6046$\pm$0.0064 & 0.6896$\pm$0.0084 \\
&  & 0.5 & 0.6589$\pm$0.0039 & 0.3868$\pm$0.0055 & 0.5900$\pm$0.0039 & 0.6806$\pm$0.0078 \\
&  & 0.7 & 0.6309$\pm$0.0032 & 0.3616$\pm$0.0044 & 0.5690$\pm$0.0034 & 0.6743$\pm$0.0109 \\
\cmidrule(lr){2-7}
& \multirow{4}{*}{AdaCoMed}
 & 0.1 & 0.6863$\pm$0.0016 & 0.4124$\pm$0.0035 & 0.4722$\pm$0.0036 & 0.5643$\pm$0.0131 \\
&  & 0.3 & 0.6649$\pm$0.0019 & 0.3861$\pm$0.0017 & 0.4583$\pm$0.0029 & 0.5472$\pm$0.0156 \\
&  & 0.5 & 0.6391$\pm$0.0030 & 0.3569$\pm$0.0027 & 0.4436$\pm$0.0050 & 0.5267$\pm$0.0213 \\
&  & 0.7 & 0.6116$\pm$0.0040 & 0.3285$\pm$0.0051 & 0.4286$\pm$0.0074 & 0.5037$\pm$0.0281 \\
\midrule
\multirow{8}{*}{MMIST-ccRCC}
& \multirow{4}{*}{MedMIX}
 & 0.1 & 0.7895$\pm$0.0167 & 0.9623$\pm$0.0039 & 0.7135$\pm$0.0201 & 0.8942$\pm$0.0142 \\
&  & 0.3 & 0.7613$\pm$0.0389 & 0.9505$\pm$0.0156 & 0.6499$\pm$0.0561 & 0.8595$\pm$0.0465 \\
&  & 0.5 & 0.6754$\pm$0.0278 & 0.9299$\pm$0.0092 & 0.5930$\pm$0.0540 & 0.8331$\pm$0.0738 \\
&  & 0.7 & 0.6708$\pm$0.0168 & 0.9296$\pm$0.0046 & 0.5982$\pm$0.0669 & 0.8265$\pm$0.0968 \\
\cmidrule(lr){2-7}
& \multirow{4}{*}{AdaCoMed}
 & 0.1 & 0.6957$\pm$0.0279 & 0.9357$\pm$0.0092 & 0.5108$\pm$0.0514 & 0.6033$\pm$0.1019 \\
&  & 0.3 & 0.6126$\pm$0.0580 & 0.9125$\pm$0.0152 & 0.4565$\pm$0.0565 & 0.5471$\pm$0.1103 \\
&  & 0.5 & 0.5558$\pm$0.0655 & 0.8920$\pm$0.0225 & 0.4236$\pm$0.0787 & 0.5289$\pm$0.1553 \\
&  & 0.7 & 0.5848$\pm$0.0944 & 0.8969$\pm$0.0264 & 0.4391$\pm$0.1030 & 0.5521$\pm$0.1897 \\
\bottomrule
\end{tabular}
}
\end{table*}

\paragraph{Efficiency metrics.}
We report $\mathrm{Perf}=(\mathrm{AUROC}+\mathrm{AUPRC}+\mathrm{mF1}+\mathrm{Acc})/4$ and $\mathrm{EffScore}=(\mathrm{Perf}/\mathrm{Perf}_0)/\mathrm{Cost}$, where $\mathrm{Cost}$ is the geometric mean of the parameter, FLOPs, and memory ratios relative to MedMIX and $\mathrm{Perf}_0$ is the MedMIX performance on the same dataset; by definition $\mathrm{EffScore}_{\text{MedMIX}}=1.000$. The efficiency accounting includes the upstream expert-feature construction modules that remain in the deployed pipeline. MeanAvg, Concat, Max, Attention, and M3Care are near-cost-matched to MedMIX as they share the same deployed expert feature constructors and differ only in the downstream fusion mechanism. On the macro-averaged row, MedMIX reaches $\mathrm{Perf}=0.780$, compared with $0.705$ for MeanAvg, $0.705$ for Concat, $0.675$ for Max, $0.693$ for Attention, $0.738$ for M3Care, $0.740$ for AdaCoMed, and $0.702$ for OneLLM.

\begin{table}[!h]
\centering
\caption{Efficiency comparison across OpenI, MIMIC-IV-MM, and MMIST-ccRCC.}
\label{tab:efficiency_all}
\tiny
\setlength{\tabcolsep}{2pt}
\renewcommand{\arraystretch}{1.05}
\resizebox{0.9\columnwidth}{!}{
\begin{tabular}{llccccc}
\toprule
Dataset & Method & Params (B) & FLOPs (G) & Peak GPU Mem. (GB) & Perf & EffScore \\
\midrule
\multirow{8}{*}{OpenI}
& MeanAvg & 2.85 & 773.74 & 14.44 & 0.759 & 0.891 \\
& Concat & 2.85 & 773.81 & 14.45 & 0.721 & 0.846 \\
& Max & 2.85 & 773.74 & 14.44 & 0.701 & 0.823 \\
& Attention & 2.85 & 773.79 & 14.44 & 0.703 & 0.825 \\
& M3Care & 2.85 & 773.79 & 14.50 & 0.814 & 0.955 \\
& AdaCoMed & 8.20 & 909.39 & 14.58 & 0.788 & 0.615 \\
& OneLLM & 8.63 & 1493.23 & 14.80 & 0.757 & 0.489 \\
& MedMIX & \textbf{2.85} & \textbf{773.74} & \textbf{14.44} & \textbf{0.851} & \textbf{1.000} \\
\midrule
\multirow{8}{*}{MIMIC-IV-MM}
& MeanAvg & 2.87 & 1751.50 & 14.21 & 0.612 & 0.959 \\
& Concat & 2.87 & 1751.59 & 14.22 & 0.613 & 0.961 \\
& Max & 2.87 & 1751.50 & 14.21 & 0.594 & 0.931 \\
& Attention & 2.87 & 1751.55 & 14.21 & 0.604 & 0.947 \\
& M3Care & 2.87 & 1751.56 & 14.26 & 0.627 & 0.982 \\
& AdaCoMed & 9.37 & 3608.26 & 15.20 & 0.621 & 0.504 \\
& OneLLM & 15.24 & 5132.74 & 18.20 & 0.536 & 0.310 \\
& MedMIX & \textbf{2.86} & \textbf{1751.50} & \textbf{14.21} & \textbf{0.637} & \textbf{1.000} \\
\midrule
\multirow{8}{*}{MMIST-ccRCC}
& MeanAvg & 3.47 & 43340.61 & 1.78 & 0.744 & 0.872 \\
& Concat & 3.47 & 43340.68 & 1.79 & 0.781 & 0.915 \\
& Max & 3.47 & 43340.61 & 1.78 & 0.730 & 0.856 \\
& Attention & 3.47 & 43340.65 & 1.78 & 0.774 & 0.907 \\
& M3Care & 3.47 & 43340.66 & 1.79 & 0.773 & 0.904 \\
& AdaCoMed & 9.88 & 41664.49 & 14.41 & 0.810 & 0.338 \\
& OneLLM & 8.64 & 59466.65 & 14.70 & 0.813 & 0.313 \\
& MedMIX & \textbf{3.46} & \textbf{43340.61} & \textbf{1.78} & \textbf{0.852} & \textbf{1.000} \\
\midrule
\multirow{8}{*}{\textbf{Average}}
& MeanAvg & 3.06 & 15288.62 & 10.14 & 0.705 & 0.903 \\
& Concat & 3.06 & 15288.69 & 10.15 & 0.705 & 0.903 \\
& Max & 3.06 & 15288.62 & 10.14 & 0.675 & 0.864 \\
& Attention & 3.06 & 15288.66 & 10.14 & 0.693 & 0.888 \\
& M3Care & 3.06 & 15288.67 & 10.18 & 0.738 & 0.944 \\
& AdaCoMed & 9.15 & 15394.05 & 14.73 & 0.740 & 0.580 \\
& OneLLM & 10.84 & 22030.87 & 15.90 & 0.702 & 0.450 \\
& MedMIX & \textbf{3.06} & \textbf{15288.62} & \textbf{10.14} & \textbf{0.780} & \textbf{1.000} \\
\bottomrule
\end{tabular}
}
\end{table}

\begin{table}[H]
\centering
\caption{Structural ablation results across OpenI, MIMIC-IV-MM, and MMIST-ccRCC.}
\label{tab:struct_ablation_three_datasets}
\setlength{\tabcolsep}{5pt}
\renewcommand{\arraystretch}{1.12}
\resizebox{0.9\textwidth}{!}{
\begin{tabular}{llcccc}
\toprule
Dataset & Setting & AUROC & AUPRC & mF1 & Acc \\
\midrule

\multirow{10}{*}{OpenI}
& MedMIX & \textbf{0.9570$\pm$0.0006} & \textbf{0.7780$\pm$0.0047} & \textbf{0.7246$\pm$0.0061} & \textbf{0.9463$\pm$0.0019} \\
& w/o distillation & 0.9456$\pm$0.0005 & \underline{0.7705$\pm$0.0049} & \underline{0.7101$\pm$0.0093} & \underline{0.9414$\pm$0.0023} \\
& w/o intra-modality fusion & \underline{0.9478$\pm$0.0015} & 0.7571$\pm$0.0046 & 0.6872$\pm$0.0052 & 0.9328$\pm$0.0058 \\
& w/o intra-modality fusion (uniform mean) & 0.9417$\pm$0.0015 & 0.7605$\pm$0.0064 & 0.6928$\pm$0.0068 & 0.9390$\pm$0.0035 \\
& w/o biomedical experts & 0.9174$\pm$0.0045 & 0.6834$\pm$0.0140 & 0.6199$\pm$0.0136 & 0.9154$\pm$0.0065 \\
& w/o general experts & 0.9349$\pm$0.0031 & 0.7130$\pm$0.0083 & 0.6623$\pm$0.0084 & 0.9281$\pm$0.0027 \\
& w/o RAG experts & 0.9475$\pm$0.0032 & 0.7700$\pm$0.0143 & 0.6922$\pm$0.0147 & 0.9336$\pm$0.0061 \\
& w/o inter-modality fusion (keep Front CXR) & 0.7752$\pm$0.0056 & 0.3599$\pm$0.0107 & 0.3480$\pm$0.0135 & 0.7286$\pm$0.0372 \\
& w/o inter-modality fusion (keep Side CXR) & 0.7464$\pm$0.0057 & 0.3106$\pm$0.0121 & 0.3332$\pm$0.0103 & 0.7508$\pm$0.0248 \\
& w/o inter-modality fusion (keep Note) & 0.9358$\pm$0.0049 & 0.7287$\pm$0.0149 & 0.6709$\pm$0.0126 & 0.9264$\pm$0.0058 \\
\midrule

\multirow{11}{*}{MIMIC-IV-MM}
& MedMIX & \textbf{0.7168$\pm$0.0019} & \textbf{0.4586$\pm$0.0019} & \textbf{0.6375$\pm$0.0035} & \textbf{0.7352$\pm$0.0106} \\
& w/o distillation & 0.7095$\pm$0.0049 & \underline{0.4420$\pm$0.0063} & 0.6205$\pm$0.0057 & 0.6894$\pm$0.0091 \\
& w/o intra-modality fusion & 0.6757$\pm$0.0017 & 0.4006$\pm$0.0022 & 0.6053$\pm$0.0020 & 0.6950$\pm$0.0167 \\
& w/o intra-modality fusion (uniform mean) & \underline{0.7106$\pm$0.0027} & 0.4402$\pm$0.0043 & \underline{0.6249$\pm$0.0054} & 0.7138$\pm$0.0119 \\
& w/o biomedical experts & 0.7080$\pm$0.0026 & 0.4414$\pm$0.0026 & 0.6231$\pm$0.0042 & 0.7096$\pm$0.0144 \\
& w/o general experts & 0.7035$\pm$0.0027 & 0.4349$\pm$0.0026 & 0.6173$\pm$0.0076 & 0.6938$\pm$0.0117 \\
& w/o RAG experts & 0.6997$\pm$0.0020 & 0.4289$\pm$0.0007 & 0.6184$\pm$0.0045 & 0.7049$\pm$0.0120 \\
& w/o inter-modality fusion (keep Front CXR) & 0.6811$\pm$0.0011 & 0.4026$\pm$0.0019 & 0.5976$\pm$0.0042 & 0.6769$\pm$0.0170 \\
& w/o inter-modality fusion (keep Lateral CXR) & 0.5305$\pm$0.0007 & 0.2685$\pm$0.0004 & 0.4326$\pm$0.0018 & \underline{0.7253$\pm$0.0271} \\
& w/o inter-modality fusion (keep EHR) & 0.5905$\pm$0.0046 & 0.3167$\pm$0.0060 & 0.5036$\pm$0.0063 & 0.7039$\pm$0.0020 \\
& w/o inter-modality fusion (keep Note) & 0.6737$\pm$0.0010 & 0.4000$\pm$0.0012 & 0.6012$\pm$0.0021 & 0.6999$\pm$0.0134 \\
\midrule

\multirow{10}{*}{MMIST-ccRCC}
& MedMIX & \textbf{0.8121$\pm$0.0068} & \textbf{0.9673$\pm$0.0019} & \underline{0.7257$\pm$0.0146} & \textbf{0.9025$\pm$0.0096} \\
& w/o distillation & 0.7615$\pm$0.0167 & 0.9526$\pm$0.0051 & 0.6407$\pm$0.0270 & 0.8099$\pm$0.0193 \\
& w/o intra-modality fusion & 0.7607$\pm$0.0122 & 0.9548$\pm$0.0050 & 0.6903$\pm$0.0189 & 0.8942$\pm$0.0033 \\
& w/o intra-modality fusion (uniform mean) & 0.7823$\pm$0.0374 & 0.9572$\pm$0.0113 & 0.7128$\pm$0.0215 & \underline{0.9008$\pm$0.0052} \\
& w/o biomedical experts & 0.6711$\pm$0.0376 & 0.9223$\pm$0.0126 & 0.5863$\pm$0.0751 & 0.8016$\pm$0.1165 \\
& w/o general experts & \underline{0.7995$\pm$0.0172} & \underline{0.9636$\pm$0.0046} & 0.7080$\pm$0.0040 & 0.8942$\pm$0.0033 \\
& w/o RAG experts & 0.7936$\pm$0.0112 & 0.9633$\pm$0.0027 & 0.7105$\pm$0.0129 & 0.8926$\pm$0.0091 \\
& w/o inter-modality fusion (keep CT) & 0.6505$\pm$0.0062 & 0.9177$\pm$0.0029 & 0.4770$\pm$0.0103 & 0.6761$\pm$0.0166 \\
& w/o inter-modality fusion (keep WSI) & 0.5056$\pm$0.0107 & 0.8744$\pm$0.0115 & 0.4727$\pm$0.0162 & 0.8364$\pm$0.0128 \\
& w/o inter-modality fusion (keep Note) & 0.7527$\pm$0.0085 & 0.9394$\pm$0.0052 & \textbf{0.7320$\pm$0.0107} & 0.8926$\pm$0.0074 \\
\bottomrule
\end{tabular}
}
\end{table}

\begin{table*}[!htbp]
\centering
\caption{Per-task performance on MIMIC-IV-MM (5 seeds, per-task threshold tuned on validation).}
\label{tab:mimic_per_task}
\scriptsize
\setlength{\tabcolsep}{3pt}
\renewcommand{\arraystretch}{1.06}
\resizebox{0.8\textwidth}{!}{
\begin{tabular}{llcccc}
\toprule
\textbf{Task} & \textbf{Method} & \textbf{AUROC} & \textbf{AUPRC} & \textbf{mF1} & \textbf{Acc} \\
\midrule
\multirow{11}{*}{hospital\_expire\_flag}
& XRay-Large       & 0.6989 & 0.2947 & 0.3547 & 0.5574 \\
& Text-Large       & 0.6645 & 0.2621 & 0.3353 & 0.5179 \\
& TimeSeries-Large & 0.5261 & 0.1699 & 0.2860 & 0.2194 \\
& MeanAvg          & 0.7151 & 0.3083 & 0.6018 & \textbf{0.7887} \\
& Concat           & 0.7144 & 0.3125 & 0.6052 & 0.7760 \\
& Max              & 0.7039 & 0.3039 & 0.5974 & 0.7406 \\
& Attention        & 0.7053 & 0.2959 & 0.6025 & 0.7522 \\
& M3Care           & \underline{0.7287} & \underline{0.3459} & \underline{0.6139} & 0.7765 \\
& AdaCoMed         & 0.7138 & 0.3105 & 0.6049 & 0.7674 \\
& OneLLM           & 0.5710 & 0.2221 & 0.5163 & 0.6847 \\
& \textbf{MedMIX}  & \textbf{0.7331} & \textbf{0.3711} & \textbf{0.6190} & \underline{0.7807} \\
\midrule
\multirow{11}{*}{longstay\_7d}
& XRay-Large       & 0.7570 & 0.5857 & 0.6303 & 0.6606 \\
& Text-Large       & 0.7885 & 0.6626 & 0.6361 & 0.6833 \\
& TimeSeries-Large & 0.5565 & 0.3828 & 0.5530 & 0.4278 \\
& MeanAvg          & 0.8076 & 0.6464 & 0.7208 & 0.7470 \\
& Concat           & 0.8075 & 0.6410 & 0.7232 & 0.7487 \\
& Max              & 0.7907 & 0.6114 & 0.7087 & 0.7029 \\
& Attention        & 0.7953 & 0.6211 & 0.7161 & 0.7195 \\
& M3Care           & \underline{0.8191} & \underline{0.6825} & \underline{0.7380} & \underline{0.7551} \\
& AdaCoMed         & 0.8119 & 0.6787 & 0.7318 & 0.7487 \\
& OneLLM           & 0.6995 & 0.5261 & 0.6313 & 0.6685 \\
& \textbf{MedMIX}  & \textbf{0.8302} & \textbf{0.6919} & \textbf{0.7463} & \textbf{0.7614} \\
\midrule
\multirow{11}{*}{readmission\_30d}
& XRay-Large       & 0.5500 & 0.2897 & 0.4000 & 0.3564 \\
& Text-Large       & 0.5799 & \underline{0.3123} & 0.4373 & 0.4048 \\
& TimeSeries-Large & 0.5054 & 0.2541 & 0.4021 & 0.2517 \\
& MeanAvg          & 0.5780 & 0.2918 & 0.5335 & 0.6156 \\
& Concat           & 0.5806 & 0.2971 & 0.5380 & 0.6277 \\
& Max              & 0.5653 & 0.2822 & 0.5206 & 0.6155 \\
& Attention        & 0.5762 & 0.2869 & 0.5232 & \underline{0.6614} \\
& M3Care           & \underline{0.5832} & 0.3055 & 0.5438 & 0.6437 \\
& AdaCoMed         & 0.5749 & \underline{0.3123} & \underline{0.5459} & 0.6503 \\
& OneLLM           & 0.5057 & 0.2551 & 0.4990 & 0.6511 \\
& \textbf{MedMIX}  & \textbf{0.5870} & \textbf{0.3127} & \textbf{0.5471} & \textbf{0.6634} \\
\bottomrule
\end{tabular}
}
\end{table*}

\section{Additional Result Figures}\label{sec:appendix_results_figures}

\begin{figure}[H]
\centering
\includegraphics[width=\textwidth]{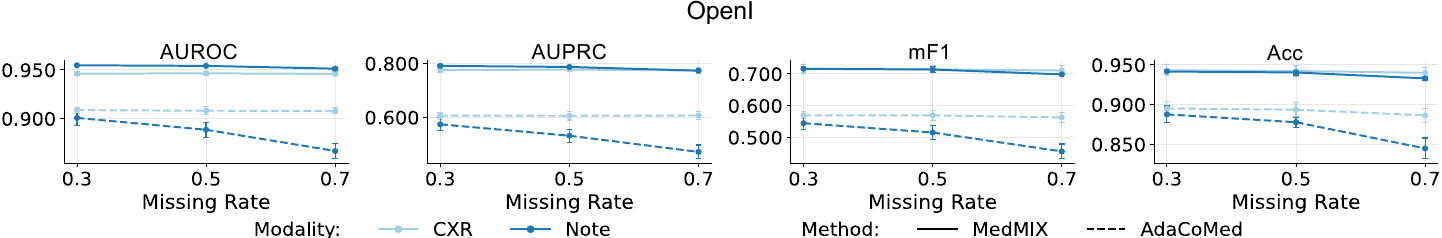}
\caption{OpenI under train-time one-modality drop. Models are trained with CXR or Note removed at rate $p\in\{0.3,0.5,0.7\}$, subplots report AUROC, AUPRC, mF1, and Acc for MedMIX versus AdaCoMed. Frontal and lateral CXR views are merged because their curves nearly overlap, full per-view results are in Appendix Table~\ref{tab:trainmissing_onemod_methods}.}
\label{fig:appendix_train_missing_onemod_openi}
\end{figure}

\begin{figure}[H]
\centering
\includegraphics[width=\textwidth]{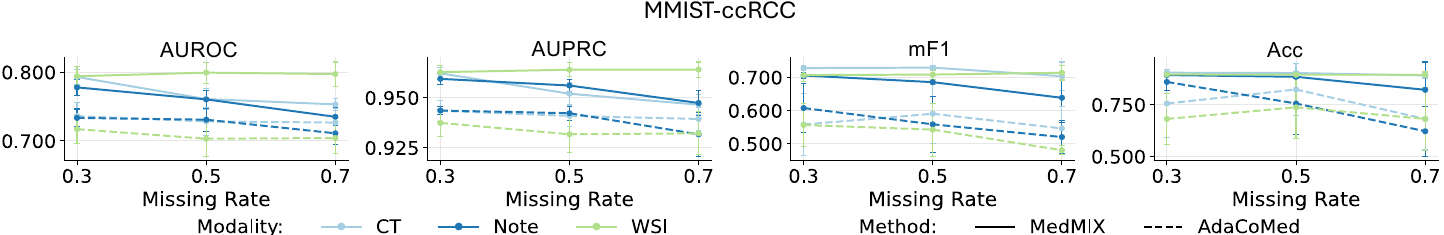}
\caption{MMIST-ccRCC under train-time one-modality drop. Models are trained with CT, WSI, or Note removed at rate $p\in\{0.3,0.5,0.7\}$, subplots report AUROC, AUPRC, mF1, and Acc for MedMIX versus AdaCoMed.}
\label{fig:appendix_train_missing_onemod_mmist}
\end{figure}

\begin{figure}[H]
\centering
\includegraphics[width=\textwidth]{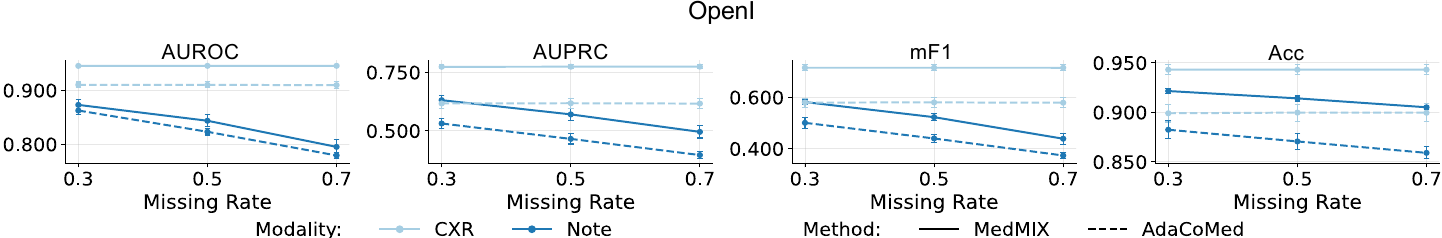}
\caption{OpenI under test-time one-modality drop. Models are evaluated with CXR or Note removed at rate $p\in\{0.3,0.5,0.7\}$, subplots report AUROC, AUPRC, mF1, and Acc for MedMIX versus AdaCoMed. Frontal and lateral CXR views are merged because their curves nearly overlap, full per-view results are in Appendix Table~\ref{tab:testmissing_onemod_medmix_vs_adacomed}.}
\label{fig:appendix_test_missing_onemod_openi}
\end{figure}

\begin{figure}[H]
\centering
\includegraphics[width=\textwidth]{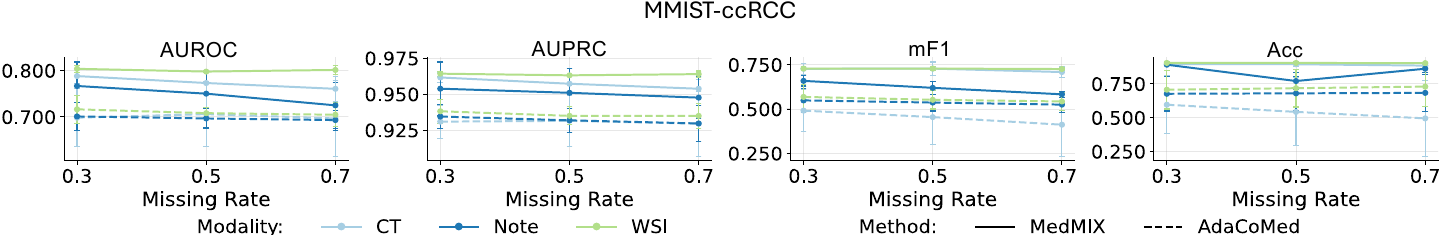}
\caption{MMIST-ccRCC under test-time one-modality drop. Models are evaluated with CT, WSI, or Note removed at rate $p\in\{0.3,0.5,0.7\}$, subplots report AUROC, AUPRC, mF1, and Acc for MedMIX versus AdaCoMed.}
\label{fig:appendix_test_missing_onemod_mmist}
\end{figure}

\begin{figure}[H]
\centering
\includegraphics[width=\textwidth]{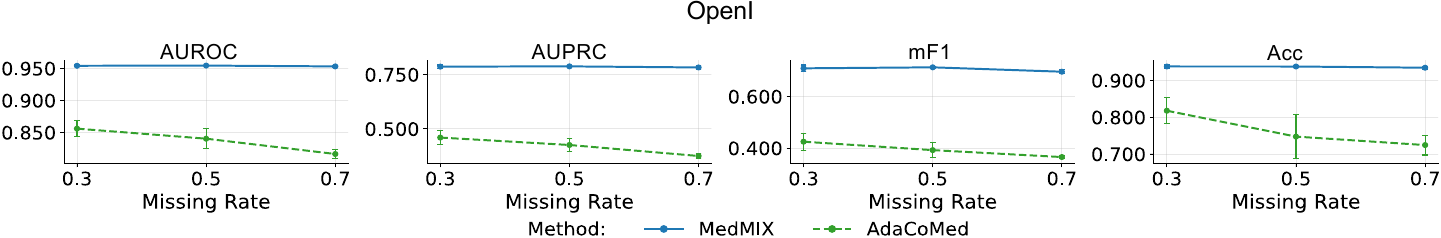}
\caption{OpenI under train-time multi-random drop. During training, each modality is independently dropped with probability $p\in\{0.1,0.3,0.5,0.7\}$, subplots report AUROC, AUPRC, mF1, and Acc for MedMIX versus AdaCoMed.}
\label{fig:appendix_train_missing_multi_openi}
\end{figure}

\begin{figure}[H]
\centering
\includegraphics[width=\textwidth]{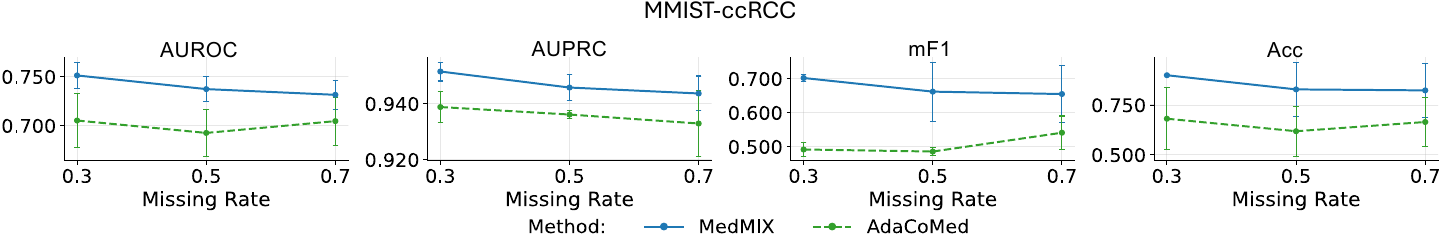}
\caption{MMIST-ccRCC under train-time multi-random drop. During training, each modality is independently dropped with probability $p\in\{0.1,0.3,0.5,0.7\}$, subplots report AUROC, AUPRC, mF1, and Acc for MedMIX versus AdaCoMed.}
\label{fig:appendix_train_missing_multi_mmist}
\end{figure}

\begin{figure}[H]
\centering
\includegraphics[width=\textwidth]{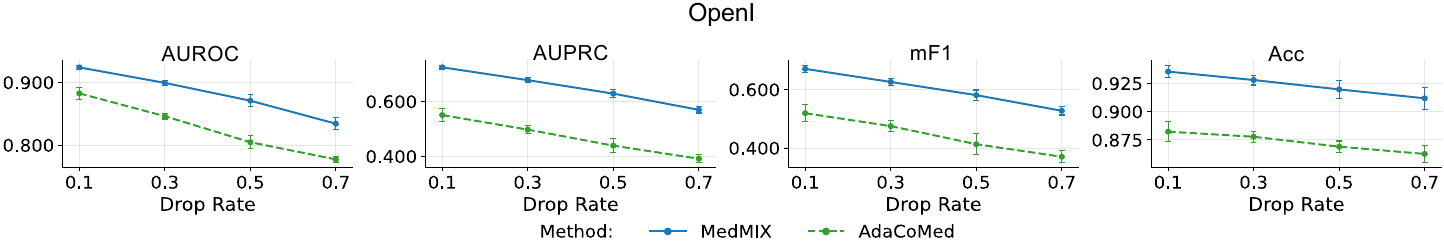}
\caption{OpenI under test-time multi-random drop. At evaluation, each modality is independently dropped with probability $r\in\{0.1,0.3,0.5,0.7\}$, subplots report AUROC, AUPRC, mF1, and Acc for MedMIX versus AdaCoMed.}
\label{fig:appendix_test_missing_multi_openi}
\end{figure}

\begin{figure}[H]
\centering
\includegraphics[width=\textwidth]{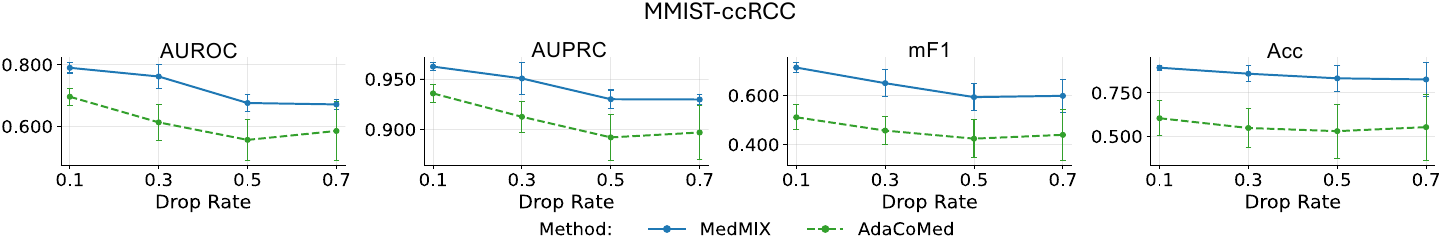}
\caption{MMIST-ccRCC under test-time multi-random drop. At evaluation, each modality is independently dropped with probability $r\in\{0.1,0.3,0.5,0.7\}$, subplots report AUROC, AUPRC, mF1, and Acc for MedMIX versus AdaCoMed.}
\label{fig:appendix_test_missing_multi_mmist}
\end{figure}





\end{document}